\documentclass[11pt,draftcls,onecolumn]{IEEEtran}

\usepackage{amsmath,amssymb,amsthm,epsfig,dsfont,algpseudocode,color,subfigure,empheq}
\DeclareMathOperator{\rank}{rank}
\DeclareMathOperator{\range}{range}
\DeclareMathOperator{\nullspace}{null}

\DeclareMathOperator{\pr}{Pr}
\DeclareMathOperator{\Beta}{B}

\newtheorem{proposition}{Proposition}
\newtheorem{lemma}{Lemma}

\newtheorem{theorem}{Theorem}
\newtheorem{definition}{Problem Statement}
\theoremstyle{remark}\newtheorem{remark}{Remark}

\begin{document}
\title{From Sparse Signals to Sparse Residuals\\ for Robust Sensing}

\author{Vassilis Kekatos,~\IEEEmembership{Member,~IEEE,} and Georgios B. Giannakis*,~\IEEEmembership{Fellow,~IEEE}%
\thanks{Part of this work was presented at the 11th IEEE Intl. Workshop on Signal Processing Advances in Wireless Communications, Marrakech, Morocco, June 2010. Work was supported by the European Community's Seventh Framework Programme (FP7/2008 under grant agreement No. 234914); and by NSF grant CCF-1016605. The authors are with the ECE Dept., University of Minnesota, Minneapolis, MN 55455, USA, Emails:\{kekatos,georgios\}@umn.edu.}%
}

\markboth{IEEE TRANSACTIONS ON SIGNAL PROCESSING (REVISED)}{Kekatos and Giannakis: From Sparse Signals to Sparse Residuals}

\maketitle 

\vspace*{-2em}
\begin{abstract}
One of the key challenges in sensor networks is the extraction of information by fusing data from a multitude of distinct, but possibly unreliable sensors. Recovering information from the maximum number of dependable sensors while specifying the unreliable ones is critical for robust sensing. This sensing task is formulated here as that of finding the maximum number of feasible subsystems of linear equations, and proved to be NP-hard. Useful links are established with compressive sampling, which aims at recovering vectors that are sparse. In contrast, the signals here are not sparse, but give rise to sparse residuals. Capitalizing on this form of sparsity, four sensing schemes with complementary strengths are developed. The first scheme is a convex relaxation of the original problem expressed as a second-order cone program (SOCP). It is shown that when the involved sensing matrices are Gaussian and the reliable measurements are sufficiently many, the SOCP can recover the optimal solution with overwhelming probability. The second scheme is obtained by replacing the initial objective function with a concave one. The third and fourth schemes are tailored for noisy sensor data. The noisy case is cast as a combinatorial problem that is subsequently surrogated by a (weighted) SOCP. Interestingly, the derived cost functions fall into the framework of robust multivariate linear regression, while an efficient block-coordinate descent algorithm is developed for their minimization. The robust sensing capabilities of all schemes are verified by simulated tests.
\end{abstract}


\section{Introduction}\label{sec:intro}
Recent advances in sensor technology have made it feasible to deploy a network of inexpensive sensors for carrying out synergistically even sophisticated inference tasks. In applications such as environmental monitoring, surveillance of critical infrastructure, agriculture, or medical imaging, the typical concept of operation involves a large and possibly heterogeneous set of sensors locally observing the signal of interest, and transmitting their measurements to a higher-layer agent (fusion center). This so-termed layered sensing apparatus entails three operational conditions:\\
(c1) Each node's measurement vector comprising either a collection of scalar observations across time, or a snapshot of different sensor readings, is typically assumed to be linearly related to the unknown variable(s). Such a \textit{linear} model can arise when the sensing system is viewed as a linear filter with known impulse response. Even when the underlying model is non-linear, the observations are approximately modeled as adhering to a (multivariate) linear regression;\\
(c2) Either because readings are costly to sense and transmit, due to delay or stationarity constraints, or simply because dimensionality reduction is invoked to cope with the ``curse of dimensionality,'' the linear model is oftentimes \emph{under-determined}, i.e., the dimension of the unknown vector is larger than that of each sensor's vector observation; and\\
(c3) Not all sensors are \emph{reliable} because failures in the sensing devices, fades of the sensor-agent communication link, physical obstruction of the scene of interest, and (un)intentional interference, all can severely deteriorate the consistency and reliability of sensor data. 

Conditions (c1)-(c3) suggest that the fusion center should not simply aggregate all sensor measurements, but instead identify and discard unreliable sensors before estimating the unknown vector based on reliable sensor data. This task is henceforth referred to as \emph{robust sensing} (RS), and provides context of the present paper. Discerning the unreliable sensors not only promises higher estimation accuracy, but also enables corrective actions to re-establish a sensor's reliability, by e.g., remotely directing the sensor to the area of interest, or, increasing its sensitivity. Even though the related problem of outlier detection in sensor networks has been studied extensively (see e.g., \cite{ZhMeHa10} for a recent survey), the RS setup and the approaches described here have not been considered before.

The first contribution of this work is to formulate the RS task as an optimization problem based on the sensor data, and show it to be NP-hard (Section \ref{sec:problem_statement}). The second one consists of two (sub)-optimum RS solvers (Section \ref{sec:RS}). The first solver is expressed as a second-order cone program (SOCP) through a convex relaxation of the original NP-hard problem. The idea of convex relaxation has been employed in the emerging area of compressive sampling (CS) \cite{ChDoSa98}, \cite{Ti96}, \cite{CaTa05}. CS asserts that a sparse vector (i.e., one having many zero entries) can be recovered with overwhelming probability as the vector with minimum $\ell_1$-norm satisfying an under-determined system of linear equations; a setup known as basis pursuit (BP) \cite{ChDoSa98}, \cite{CaTa05}, \cite{Tro06}. CS has been generalized to block-sparse signals, where the unknown vector comprises predetermined subsets of variables that are (non) zero as a group \cite{StPaHa09}, \cite{St09}, \cite{ElMi09}, \cite{BaCeDuHe10}. Block sparsity emerges also in the RS formulation herein, not in the unknown vector though, but in the per-sensor residual error vectors. The relation between recovering block-sparse signals and the developed RS solver nicely generalizes the equivalence of BP with $\ell_1$-error regression from the scalar to the vector case. As an alternative to convex relaxation, the $\ell_0$-(pseudo)norm of the wanted vector can be replaced by a concave approximation to further promote sparsity \cite{Fa02}, \cite{CaWeBo08}. This constitutes the second RS solver, which surrogates the original objective by a concave function, and minimizes it through a sequence of weighted SOCPs.

The third contribution consists in analyzing the performance (identifiability) of the convex relaxation approach to recover the unknown vector, and successfully select the reliable sensors in the noise-free case (Section \ref{sec:performance}). The analysis hinges on a set of necessary and sufficient conditions on the involved matrix range space, which appear also in the context of \cite{StPaHa09}. Here a lower bound expressed in closed form is established on the probability of success when the design matrix is drawn from the Gaussian ensemble; see also \cite{ReFaPa07}. It is shown that whenever there is sufficient majority of reliable sensors and quantifiably enough per-sensor measurements, the solution of the SOCP is exact with overwhelming probability.

In real-world applications, sensor readings are contaminated by additive noise due to quantization, communication noise, and/or unmodeled dynamics. Besides identifiability, the aforementioned schemes are thus appropriate only for the high signal-to-noise ratio (SNR) regime. When the sparse vector in CS is observed in noise, its recovery is based on methods such as the Lasso \cite{Ti96}, or the group Lasso for vectors that are block-sparse \cite{YuLi06}. Different from CS, the approach here views the unreliable sensors as outliers, thus placing the sensing in the presence of noise (RSN) task under a robust multivariate linear regression framework \cite{AeWi05}, \cite{BaRaWu92}. The fourth contribution of this work (Section \ref{sec:RSN}) is initially formulating RSN as a combinatorial optimization problem that is subsequently surrogated by a convex approximation. Interestingly, the novel cost function turns out to be a block version of Huber's function \cite{HuRo09}. The resultant optimization problem is transformed to a group Lasso-type SOCP, and a computationally attractive block-coordinate descent algorithm is developed. An alternative RSN solver is also offered after replacing the previously derived convex problem with a non-convex one. The simulated tests presented in Section \ref{sec:simulations} corroborate the proposed schemes, and the paper is concluded in Section \ref{sec:conclusions}.

\emph{Notation:} Lowercase (upper-case) boldface letters are reserved for column vectors (matrices), and calligraphic letters for sets; $(\cdot)^T$ denotes transposition; $\mathcal{N}(\mathbf{m},\mathbf{\Sigma})$ stands for the multivariate Gaussian probability density with mean $\mathbf{m}$ and covariance matrix  $\mathbf{\Sigma}$, while $\mathbb{E}[\cdot]$ denotes the expectation operator. The notation $\|\mathbf{x}\|_p:=\left(\sum_{i=1}^n |x_i|^p\right)^{1/p}$ for $p=1(2)$ stands for the $\ell_1$($\ell_2$)-norm in $\mathbb{R}^n$, and $\|\mathbf{x}\|_0$ the $\ell_0$-(pseudo)norm which equals the number of nonzero entries of $\mathbf{x}$.

\section{Preliminaries and Problem Statement}\label{sec:problem_statement}
Consider an agent, e.g., an unmanned aerial vehicle, collecting data vectors $\{\mathbf{b}_i\}_{i=1}^k$ of size $m_i\times 1$, and corresponding $m_i\times n$ regression matrices $\{\mathbf{A}_i\}_{i=1}^k$ from $k$ sensors. The goal is to find an unknown vector $\mathbf{x}\in \mathbb{R}^n$, possibly satisfying the linear subsystems of equations $\mathbf{b}_i=\mathbf{A}_i\mathbf{x}$ for some $i\in \{1,\ldots,k\}$. This goal is challenging since the unknown vector $\mathbf{x}$ satisfies only an \emph{unknown} subset of sensors. The RS problem can be compactly stated as follows.

\begin{definition}[Robust sensing (RS)]\label{def:RS}
Given $k$ vector-matrix pairs, $\left\{\mathbf{b}_i, \mathbf{A}_i \right\}_{i=1}^{k}$, where $\mathbf{b}_i\in\mathbb{R}^{m_i}$ and $\mathbf{A}_i\in\mathbb{R}^{m_i\times n}$, find a vector $\mathbf{x}\in\mathbb{R}^n$ that maximizes the number of feasible linear subsystems $\{\mathbf{b}_i=\mathbf{A}_i\mathbf{x}\}$.
\end{definition}

\begin{figure}[!ht]
\centering
\includegraphics[width=0.6\linewidth]{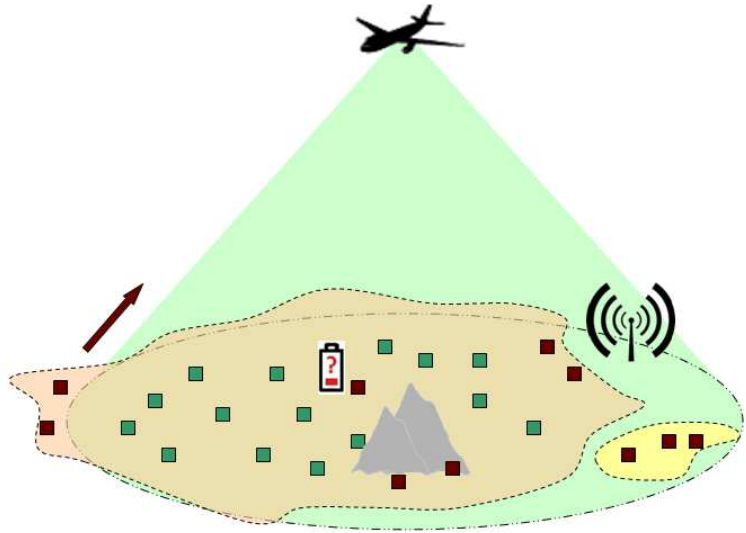}
\vspace*{-1em}
\caption{A wireless sensor network linked with a fusion center. (Un)reliable sensors are color coded as (red) green.}
\vspace*{-1em}
\label{fig:sn}
\end{figure}

Vector $\mathbf{x}$ could model a scene (lexicographically ordered image) of interest viewed by multiple and possibly heterogeneous, e.g., Infrared, SAR, or, Lidar imaging systems. Matrices $\mathbf{A}_i$ may capture variable fields of view, different perspectives and resolutions in some (e.g., wavelet) domain, or, calibration parameters of the respective sensors. Alternatively, in
an environmental monitoring application, $\mathbf{x}$ could represent the unknown parameters of a chemical/biological compound diffusion field described by the Green's function captured by
the matrices $\{\mathbf{A}_i\}_{i=1}^k$, and measured by a wireless sensor network deployed over a region of interest. In such sensing applications, a sensor may reckoned unreliable or irrelevant due to obstruction, fading propagation effects, device failures, jamming, or, even because it collects data corresponding to an irrelevant $\mathbf{x}'\neq \mathbf{x}$; see Fig.~\ref{fig:sn}.

%%% \label{fig:sn}

The RS task is different for over- and under-determined linear subsystems. Assume that all $\mathbf{A}_i$'s are full rank, i.e., $\rank(\mathbf{A}_i)=\min\{m_i,n\}$ for all $i$.\footnote{This is without loss of generality (w.l.o.g.), because every sensor with $\rank(\mathbf{A}_i)<\min\{m_i,n\}$ will be either infeasible, or, it can be transformed to an under-determined subsystem with full row rank.} Then, suppose that the $i$-th linear subsystem is over-determined $(m_i>n)$. This subsystem is either infeasible and can be ignored, or, it admits a unique solution $\check{\mathbf{x}}_i$. In the latter case, it can be easily checked whether $\check{\mathbf{x}}_i$ satisfies any other subsystem. The solution $\check{\mathbf{x}}_i$ together with the total number of subsystems it satisfies are retained, and the method proceeds similarly with all other over-determined subsystems. However, checking the under-determined subsystems $(m_i<n)$ is more challenging, since each one of them admits infinitely many solutions. Recognizing that over-determined subsystems can be easily handled, this paper focuses on the RS task when $m_i<n$ for all $i$. Note that under-determinacy may arise naturally because of stringent power, bandwidth, delay, or stationarity constraints. Given that the $\mathbf{b}_i$'s ($\mathbf{A}_i$'s) can be padded with zero entries (rows) to match the dimension $\max_i{m_i}$, it will be henceforth assumed w.l.o.g. $m_i=m<n$ for all $i$.

Before proceeding, it is useful to introduce some parameters. The set of all subsystem indices is denoted by $\mathcal{I}:=\{1,\ldots,k\}$, whereas the pair $(\mathcal{S},\bar{\mathcal{S}})$ denotes a partition of $\mathcal{I}$ into the subset $\mathcal{S}$ and its complement $\bar{\mathcal{S}}$ $\left(\mathcal{S}\cup\bar{\mathcal{S}} =\mathcal{I}, ~\mathcal{S}\cap\bar{\mathcal{S}} =\emptyset\right)$. Consider now the $|\mathcal{S}|m\times n$ matrix $\mathbf{A}_{\mathcal{S}}$ constructed by concatenating the matrices $\{\mathbf{A}_i\}_{i\in\mathcal{S}}$, and likewise for the vector $\mathbf{b}_{\mathcal{S}}$. The aggregate regression matrix and data vector are defined as $\mathbf{A}^T:= \left[\mathbf{A}_1^T ~\ldots~\mathbf{A}_k^T\right]$ and $\mathbf{b}^T:=\left[\mathbf{b}_1^T ~\ldots~\mathbf{b}_k^T\right]$, respectively.

Upon introducing an auxiliary vector $\mathbf{t}\in \mathbb{R}^k$, the RS problem can be rigorously posed as 
\begin{empheq}[box=\fbox]{align}\label{eq:P0}
\min_{\mathbf{x},\mathbf{t}} ~&~ \|\mathbf{t}\|_0\tag{$P_0$}\\
\textrm{s.t.} ~&~ \|\mathbf{b}_i-\mathbf{A}_i\mathbf{x}\|_2\leq t_i,~i=1,\ldots,k\nonumber.
\end{empheq}
If the $i$-th subsystem is deemed feasible, then $t_i=0$; otherwise, $t_i$ is strictly positive and the cost $\|\mathbf{t}\|_0$ increases. In a nutshell, \eqref{eq:P0} minimizes the number of infeasible linear subsystems, and hence solves RS. Note also that the constraints are satisfied as equalities at the optimum. Thus, if the optimum $\mathbf{x}$ is given, the optimum $\mathbf{t}$ is readily available. This implies that the solution pair $(\mathbf{x},\mathbf{t})$ is identified solely by $\mathbf{x}$, which will be henceforth called the \textit{solution} of \eqref{eq:P0}.

Even though the constraints in \eqref{eq:P0} are convex, the problem is non-convex. A greedy approach to solving it would be to assume there are $s$ feasible subsystems, and let $s$ range from $k$ down to 1. For each value of $s$, one can check feasibility of the linear systems $\mathbf{b}_{\mathcal{S}} = \mathbf{A}_{\mathcal{S}}\mathbf{x}$ for each of the $\binom{k}{s}$ subsets $\mathcal{S}\subset \mathcal{I}$ having cardinality $|\mathcal{S}|=s$, until a feasible subset is found. But this approach incurs combinatorial complexity, and can be computationally feasible only for small-size problems. In fact, it is not difficult to establish the following result.

\begin{proposition}\label{pr:NP-hard}
The RS problem is NP-hard.
\end{proposition}
\begin{IEEEproof}
Consider first the following problem of maximizing the number of consistent linear equations (MCLE): \emph{``Given a system of linear equations $\mathbf{C} \mathbf{x} = \mathbf{d}$, where $\mathbf{C}\in \mathbb{R}^{k\times n}$ and $\mathbf{d}\in \mathbb{R}^{k}$, find a vector $\mathbf{x}\in \mathbb{R}^n$ satisfying as many equations as possible.''} The MCLE problem is known to be NP-hard \cite[Th.~1]{AmKa95}. Consider an instance of the MCLE problem. Choose an integer $m\geq 2$ and define the instance of RS with parameters $(\mathbf{b},\mathbf{A})$ selected as $b_{(i-1)m+1}=d_{i}$ and $A_{(i-1)m+1,j}=C_{i,j}$ for $i=1,\ldots,k$ and $j=1,\ldots,n$; and 0 for their remaining entries. Solving an MCLE problem is hence equivalent to solving an instance of an RS. This simple reduction of MCLE to RS establishes the proposition.
\end{IEEEproof}

In search of sub-optimum yet computationally affordable solvers of \eqref{eq:P0}, one could adopt the least-squares (LS) approach, which amounts to
\begin{equation}\label{eq:LS}
\min_{\mathbf{x}} \|\mathbf{b}-\mathbf{A}\mathbf{x}\|_2^2.
\end{equation}
Alternatively, one could consider minimizing the $\ell_1$-norm of the error, namely
\begin{equation}\label{eq:L1}
\min_{\mathbf{x}}\|\mathbf{b}- \mathbf{A}\mathbf{x}\|_1.
\end{equation}
Unfortunately, both approaches handle separately every linear
equation, and thus ignore the underlying per-sensor linear subsystem. In addition, they cannot reliably identify the unreliable sensors.

\section{RS Solvers}\label{sec:RS}

\subsection{A Convex Relaxation Solver}\label{subsec:RS_convex}
It is known that if the infinity norm satisfies $\|\mathbf{t}\|_{\infty}:=\max_i |t_i|\leq 1$, then the $\ell_1$-norm $\|\mathbf{t}\|_1$ is the convex envelope (the largest convex under-approximant) of $\|\mathbf{t}\|_0$; see e.g., \cite[p. 119]{BoVa04}. This property is used also in CS \cite{Tro06}, and prompts one to relax the NP-hard problem \eqref{eq:P0} to
\begin{align}\label{eq:P1con}
\min_{\mathbf{x},\mathbf{t}}~&~\|\mathbf{t}\|_1\\
\textrm{s.t.} ~&~ \|\mathbf{b}_i- \mathbf{A}_i \mathbf{x}\|_2\leq
t_i,~i=1,\ldots,k\nonumber.
\end{align}
Note though that $\mathbf{x}$ here does not have to be sparse. The problem in \eqref{eq:P1con} is an SOCP and can be efficiently solved by several existing algorithms \cite{BoVa04}. Invoking the implicit constraint $\mathbf{t}\geq \mathbf{0}$ and the definition of the $\ell_1$-norm $\|\mathbf{t}\|_1:=\sum_{i=1}^k|t_i|$, the problem \eqref{eq:P1con} is equivalent to 
\begin{empheq}[box=\fbox]{align}\tag{$P_1$}\label{eq:P1}
 \min_{\mathbf{x}}\sum_{i=1}^k\|\mathbf{b}_i-\mathbf{A}_i\mathbf{x}\|_2
\end{empheq}
which is still an SOCP, albeit unconstrained.

The cost in \eqref{eq:P1} is the sum of the $\ell_2$-norms of the residual vectors associated with the linear subsystems, which is continuous, but
not differentiable. In the optimization circles, \eqref{eq:P1} is known as the minimization of the sum of (Euclidean) norms problem \cite[Sec. 6.4]{BoVa04}. It emerges also when solving problems related to Steiner trees, optimal location, and image restoration model constraints; see e.g., \cite[Sec. 2.2]{LoVaLeBo98}, and references therein. Algorithmically, \eqref{eq:P1} is tackled either by generic SOCP solvers, or, by interior-point algorithms customized to its specific form \cite{LoVaLeBo98}.

Having relaxed the RS problem \eqref{eq:P0} to its closest convex approximation \eqref{eq:P1} which is tractable, it is of interest to reflect on various links and interpretations that \eqref{eq:P1} can afford, postponing its performance analysis to Section \ref{sec:performance}.

\begin{remark}[\eqref{eq:P1} versus LS]
Clearly, the LS problem in \eqref{eq:LS} can be rewritten as
\begin{equation*} \min_{\mathbf{x},\mathbf{t}}\left\{\|\mathbf{t}\|_2:
~\|\mathbf{b}_i-\mathbf{A}_i\mathbf{x}\|_2\leq
t_i,~i=1,\ldots,k\right\}
\end{equation*}
which is again a convex approximation of \eqref{eq:P0}, though, as mentioned earlier, not the closest one.
\end{remark}

%\begin{remark}[\eqref{eq:P1} as regularized LS]
%Consider squaring the cost of \eqref{eq:P1} to obtain
%\begin{equation}\label{eq:P1_squared}
%\min_{\mathbf{x}}\|\mathbf{b}-\mathbf{A}\mathbf{x}\|_2^2+2\sum_{i=1}^k
%\sum_{j=i+1}^k \|\mathbf{b}_i-\mathbf{A}_i\mathbf{x}\|_2\|\mathbf{b}_j
%-\mathbf{A}_j\mathbf{x}\|_2.
%\end{equation}
%The cost in \eqref{eq:P1_squared} reveals that \eqref{eq:P1} actually minimizes the
%conventional LS error \eqref{eq:LS} regularized by the sum of all the pairwise products of the per-sensor $\ell_2$ residual error norms.
%\end{remark}

\begin{remark}[\eqref{eq:P1} versus block-sparse signal reconstruction] \label{re:MSNvsBSSR}
To establish this connection, assume that $\nullspace(\mathbf{A}^T)$ is non-empty. Let $\mathbf{r}_i:= \mathbf{b}_i-\mathbf{A}_i \mathbf{x}$ denote the residual error vectors, and $\mathbf{r}^T:= [\mathbf{r}_1^T~\cdots~\mathbf{r}_k^T]$. Upon defining matrix $\mathbf{C}$ such that its null space is spanned by $\range(\mathbf{A})$, i.e., $\mathbf{CA}=\mathbf{0}$, and $\mathbf{d}:=\mathbf{C}\mathbf{b}$, the problem \eqref{eq:P1} can be rendered equivalent to
\begin{subequations}\label{eq:BBP}
\begin{align}
\min_{\mathbf{r}} ~&~ \sum_{i=1}^k \|\mathbf{r}_i\|_2\label{eq:BBPa}\\
\textrm{s.t.} ~&~ \mathbf{C}\mathbf{r}=\mathbf{d}\label{eq:BBPb}
\end{align}
\end{subequations}
which emerges when reconstructing a \emph{block-sparse} vector $\mathbf{r}$ satisfying the under-determined system in \eqref{eq:BBPb} \cite{StPaHa09}, \cite{St09}, \cite{ElMi09}, \cite{BaCeDuHe10}. To establish the equivalence, write \eqref{eq:P1} as $\min_{\mathbf{r}} \sum_{i=1}^k \|\mathbf{r}_i\|_2$ subject to $\mathbf{r}=\mathbf{b}-\mathbf{A}\mathbf{x}$. Premultiplying both sides of the last equality by $\mathbf{C}$, one arrives at \eqref{eq:BBP}. The same equality couples the minimizers of the two problems: if $\mathbf{r}_0$ solves \eqref{eq:BBP} and $\mathbf{A}^{\dag}$ is the pseudo-inverse of $\mathbf{A}$, then $\mathbf{A}^{\dag}(\mathbf{b}-\mathbf{r}_0)$ solves \eqref{eq:P1}. The optimization in \eqref{eq:BBP} relies on the prior information that $\mathbf{r}$ is block sparse. For the RS problem, the vector of interest $\mathbf{x}$ is not (block) sparse; but the residual error vector is block sparse.
\end{remark}

\begin{remark}[\eqref{eq:P1} versus $\ell_1$-error regression]\label{re:MSNvsL1}
In the degenerate case $m=1$, where every subsystem reduces to a single equation, \eqref{eq:P1} reduces to the $\ell_1$-error minimization problem \eqref{eq:L1}, which is known to be robust to outliers \cite[Ch.~4]{MaMaYo06}, \cite{BoVa04}, \cite{CaTa05}. Under the conditions stated in Remark \ref{re:MSNvsBSSR}, the unconstrained $\ell_1$-error regression problem is equivalent to the constrained optimization (cf. \eqref{eq:BBP})
\begin{align}\label{eq:BP}
\min_{\mathbf{r}} ~&~ \|\mathbf{r}\|_1\\
\textrm{s.t.} ~&~\mathbf{C}\mathbf{r}=\mathbf{d}.\nonumber
\end{align}
The problem in \eqref{eq:BP} is widely known in the CS literature as basis pursuit (BP); for a thorough treatment on this pair of problems see also \cite{CaTa05}.
\end{remark}

\subsection{A Concave Surrogate for RS}\label{subsec:RS_concave}
Instead of substituting the cost $\|\mathbf{t}\|_0$ of \eqref{eq:P0} by its closest convex approximation, namely $\|\mathbf{t}\|_1$, letting the surrogate function be non-convex can yield tighter approximations. For example, the $\ell_0$-norm of a vector $\mathbf{x}\in \mathbb{R}^n$ was surrogated in \cite{CaWeBo08} by the logarithm of the geometric mean of its elements, or, by $\sum_{i=1}^n\log |x_i|$. In rank minimization problems, apart from the nuclear norm relaxation, minimizing the logarithm of the determinant of the unknown matrix has been proposed as an alternative surrogate; see \cite[Sec. 5.2]{Fa02}. Building on this line of thought, consider surrogating \eqref{eq:P0} by 
\begin{empheq}[box=\fbox]{align}\tag{$P_2$}\label{eq:P2}
\min_{\mathbf{x},\mathbf{t}} ~&~ 
\sum_{i=1}^{k}\log\left(t_i+\delta\right)\\
\textrm{s.t.} ~&~ \|\mathbf{b}_i-\mathbf{A}_i\mathbf{x}\|_2\leq t_i,~i=1,\ldots,k\nonumber
\end{empheq}
where $\delta$ is a sufficiently small but strictly positive constant preventing the cost from tending to $-\infty$. The cost in \eqref{eq:P2} is concave, but since it is smooth wrt $\mathbf{t}\in\mathbb{R}_+^k$, iterative linearization may be utilized to obtain a local minimum \cite{Fa02}, \cite{CaWeBo08}. Specifically, let $(\mathbf{x}^{(l)}, \mathbf{t}^{(l)})$ denote a tentative solution at the $l$-th iteration. Due to the concavity of the logarithm, the first-order approximation of $\log\left(t_i+\delta\right)$ around $t_i^{(l-1)}+\delta$ yields
\begin{equation}\label{eq:MM}
\log\left(t_i+\delta\right) \leq \log\left(t_i^{(0)}+\delta\right) + \frac{1}{t_i^{(0)}+\delta}\left(t_i-t_i^{(0)}\right).
\end{equation}
Thinking along the majorization-minimization approach \cite{LaHuYa00}, one can instead of minimizing the original cost on the left-hand side, minimize the majorizing cost on the right-hand side of \eqref{eq:MM}, and iterate. Specifically, the minimization in \eqref{eq:P2} can be iteratively driven to a local minimum \cite{Fa02} as
\begin{align*}
\left(\mathbf{x}^{(l)},\mathbf{t}^{(l)}\right):=
\arg \min_{\mathbf{x},\mathbf{t}} \left\{ \sum_{i=1}^{k}\frac{t_i}{t_i^{(l-1)}+\delta}:~\|\mathbf{b}_i-\mathbf{A}_i\mathbf{x}\|_2\leq t_i,~i=1,\ldots,k\right\}\nonumber
\end{align*}
or equivalently,
\begin{align}\label{eq:P2a}
\mathbf{x}^{(l)}:=
\arg \min_{\mathbf{x}}
\sum_{i=1}^{k}\frac{\|\mathbf{b}_i-\mathbf{A}_i\mathbf{x}\|_2}{\|\mathbf{b}_i-\mathbf{A}_i\mathbf{x}^{(l-1)}\|_2+\delta}.
\end{align}
The iterative scheme can be terminated as soon as the relative error $\|\mathbf{x}^{(l)}-\mathbf{x}^{(l-1)}\|_2 / \|\mathbf{x}^{(l-1)}\|_2$ becomes smaller than some $\epsilon$ chosen equal to say $10^{-6}$. The cost in \eqref{eq:P2a} has the form of a weighted version of \eqref{eq:P1}, where each of the error norms is weighted by $w_i^{(l)}=\left(\|\mathbf{b}_i- \mathbf{A}_i\mathbf{x}^{(l-1)}\|_2 + \delta\right)^{-1}$. When the residual error of a subsystem is small, then the error of this system is weighted more during the minimization of the next iteration. A good initialization point for the iteration in \eqref{eq:P2a} is the solution of \eqref{eq:P1} that is equivalent to one iteration of \eqref{eq:P2a} with all weights chosen equal. The simulated tests in Section \ref{sec:simulations} will indicate that \eqref{eq:P2a} can provide higher probability of identifying reliable sensors than \eqref{eq:P1}.

\section{Uniqueness and Identifiability}\label{sec:performance}
Let $s$ denote the minimum cost of \eqref{eq:P0}. Then, there exists at least one unknown $\mathbf{x}_0\in \mathbb{R}^n$ such that $\mathbf{b}_{\mathcal{S}_0} = \mathbf{A}_{\mathcal{S}_0}\mathbf{x}_0$ for an unknown subset of sensors $\mathcal{S}_0$ with $|\mathcal{S}_0|= s$. The sensors in $\mathcal{S}_0$ will be referred to as \textit{reliable} or \textit{consistent} with respect to (w.r.t.) $\mathbf{x}_0$. Also, let $\beta:=s/k$ denote the number of consistent sensors over the total number of sensors; and $\gamma:=n/(km)$ the ratio of the size of the unknown vector over the total number of measurements.

Whether \eqref{eq:P0} has a unique minimizer, and hence an underlying $\mathbf{x}_0$ can be uniquely recovered by \eqref{eq:P0}, is considered next. The first thing to note at the outset is that when the consistent sensors w.r.t. $\mathbf{x}_0$ are outnumbered by the unreliable ones, uniquely recovering $\mathbf{x}_0$ is not guaranteed. This is because with $s\leq k/2$, there may exist an $\mathbf{x}_1\neq \mathbf{x}_0$ and an $\mathcal{S}_1\subset \mathcal{I}$ with $|\mathcal{S}_1|=|\mathcal{S}_0|= s$ and $\mathcal{S}_1 \cap \mathcal{S}_0=\emptyset$ such that $\mathbf{b}_{\mathcal{S}_1} = \mathbf{A}_{\mathcal{S}_1}\mathbf{x}_1$; thus, $\mathbf{x}_0$ and $\mathbf{x}_1$ are both minimizers of \eqref{eq:P0}. It is henceforth assumed that $s>k/2$ or $\beta\in (1/2,1]$. Under this assumption, uniqueness of the \eqref{eq:P0} minimizer is further characterized in the following lemma.

\begin{lemma}\label{le:unique}
Let vector $\mathbf{x}_0$ be a minimizer of \eqref{eq:P0} satisfying $s>k/2$ out of the $k$ subsystems. This minimizer is unique if and only if 
\begin{equation}\label{eq:unique}
\rank(\mathbf{A}_{\mathcal{S}_c})=n
\end{equation}
for every $\mathcal{S}_c\subset \mathcal{I}$ with cardinality $|\mathcal{S}_c|=2s-k$.
\end{lemma}

\begin{IEEEproof} 
Vector $\mathbf{x}_0$ is not the unique minimizer of \eqref{eq:P0} if and only if there exists at least one $\mathbf{x}_1\neq \mathbf{x}_0$ such that $\mathbf{b}_{\mathcal{S}_1} = \mathbf{A}_{\mathcal{S}_1}\mathbf{x}_1$ for an $\mathcal{S}_{1}\subset \mathcal{I}$ with $|\mathcal{S}_1|=|\mathcal{S}_0|=s$. Given that $s>k/2$, the two subsets cannot be disjoint; hence, they must have a non-empty intersection $\mathcal{S}_{c}:=\mathcal{S}_0 \cap \mathcal{S}_1$ with cardinality $2s-k\leq |\mathcal{S}_{c}|\leq s$. The subsystems belonging to $\mathcal{S}_c$ are satisfied by both solutions; that is, $\mathbf{b}_{\mathcal{S}_c} = \mathbf{A}_{\mathcal{S}_c}\mathbf{x}_{0} = \mathbf{A}_{\mathcal{S}_c}\mathbf{x}_{1}$, which is equivalent to the existence of a nonzero $\mathbf{z}\in\mathbb{R}^n$ such that $\mathbf{A}_{\mathcal{S}_c}\mathbf{z} = \mathbf{0}$ or $\rank(\mathbf{A}_{\mathcal{S}_c})<n$. Multiple minimizers of \eqref{eq:P0} can thus be avoided if and only if $\rank(\mathbf{A}_{\mathcal{S}_c})=n$. Note that whenever $\rank(\mathbf{A}_{\mathcal{S}_c})= n$ for every $\mathcal{S}_c$ with $|\mathcal{S}_c|=2s-k$, it holds for every $\mathcal{S}_c$ of larger cardinality as well.
\end{IEEEproof}

Lemma \ref{le:unique} reveals two interesting points on uniquely recovering $\mathbf{x}_0$ by \eqref{eq:P0}. First, the reliable sensors should not only outnumber the unreliable ones, i.e., $\beta>1/2$; condition \eqref{eq:unique} implies additionally that $(2s-k)m\geq n$, or $\beta\geq \left(\gamma +1\right)/2$. Second, because $\beta\leq 1$, the inequality $\beta\geq \left(\gamma +1\right)/2$ implies $\gamma\leq 1$ or $km\geq n$, requiring the total number of equations to be at least equal to the number of unknowns.

Uniqueness of the \eqref{eq:P0} minimizer is also implied by the conditions stated in the next lemma. These conditions will be used in the next subsection.

\begin{lemma}\label{le:unique-exact}
If for any nonzero $\mathbf{v}\in \range(\mathbf{A})$ and any partition $(\mathcal{S},\bar{\mathcal{S}})$ of $\mathcal{I}$ with $|\mathcal{S}|=s>k/2$ it holds that
\begin{equation}\label{eq:rsconditions2}
\sum_{i\in \mathcal{S}}\|\mathbf{v}_i\|_2 > \sum_{i\in \bar{\mathcal{S}}}\|\mathbf{v}_i\|_2
\end{equation}
where $\mathbf{v}_i$ is the $i$-th $m\times 1$ block subvector of $\mathbf{v}$, then \eqref{eq:unique} is satisfied.
\end{lemma}

\begin{IEEEproof}
Arguing by contradiction, suppose that \eqref{eq:rsconditions2} holds, whereas \eqref{eq:unique} does not hold; or, in other words there exists an $\mathcal{S}_c\subset \mathcal{I}$ with $|\mathcal{S}_c|=2s-k\leq s$ and $\rank(\mathbf{A}_{\mathcal{S}_c})<n$. Consequently, there exists a nonzero vector $\mathbf{u}\in \mathbb{R}^n$ such that $\mathbf{A}_{\mathcal{S}_c}\mathbf{u}=\mathbf{0}$. Next, partition $\mathcal{I}$ into three collectively exhaustive and mutually exclusive subsets $\mathcal{S}_c$, $\mathcal{S}_1$, and $\mathcal{S}_2$, with $|\mathcal{S}_1|=|\mathcal{S}_2|=k-s$. Define also $\mathbf{v}:=\mathbf{A}\mathbf{u}$ for which $\mathbf{v}_{\mathcal{S}_c}=\mathbf{0}$ by the definition of $\mathbf{u}$. 

Consider first \eqref{eq:rsconditions2} with $\mathcal{S}=\mathcal{S}_c\cup \mathcal{S}_1$ and $\bar{\mathcal{S}}=\mathcal{S}_2$, to deduce that
\begin{equation*}
\sum_{i\in \mathcal{S}_1}\|\mathbf{v}_i\|_2 > \sum_{i\in \mathcal{S}_2}\|\mathbf{v}_i\|_2.
\end{equation*}
Apply \eqref{eq:rsconditions2} again for $\mathcal{S}=\mathcal{S}_c\cup \mathcal{S}_2$ and $\bar{\mathcal{S}}=\mathcal{S}_1$, to arrive at
\begin{equation*}
\sum_{i\in \mathcal{S}_2}\|\mathbf{v}_i\|_2 > \sum_{i\in \mathcal{S}_1}\|\mathbf{v}_i\|_2
\end{equation*}
which clearly contradicts the previous inequality and completes the proof.
\end{IEEEproof}

Having introduced the convex relaxation \eqref{eq:P1} of \eqref{eq:P0}, the next critical question is whether the solution of the former coincides with the solution of the latter. Even though the NP-hardness of \eqref{eq:P0} forejudges that this cannot hold in general, the ensuing results show that for random Gaussian matrices $\mathbf{A}$ and under reasonable assumptions on the problem dimensions, equivalence of \eqref{eq:P1} and \eqref{eq:P0} occurs with probability exponentially decaying in $n$. The analysis starts by characterizing this equivalence using a set of necessary and sufficient conditions.

\subsection{Necessary and Sufficient Conditions}\label{susec:conditions}
The conditions under which the convex optimization problem \eqref{eq:P1} yields the same solution as the NP-hard problem \eqref{eq:P0} are provided in the following theorem. Using the equivalence between \eqref{eq:P1} and \eqref{eq:BBP} under the conditions of Remark \ref{re:MSNvsBSSR}, this theorem is related to \cite[Th.~2]{StPaHa09}, which in turn, generalizes results from \cite{DoHu01} to the block-sparse signal case.

\begin{theorem}[Range space conditions]\label{th:rsconditions}
Every $\mathbf{x}_0$ minimizing \eqref{eq:P0} by satisfying $s>k/2$ out of the $k$ subsystems is the unique minimizer of \eqref{eq:P1} if and only if 
\begin{equation}\label{eq:rsconditions}
\sum_{i\in \mathcal{S}}\|\mathbf{v}_i\|_2 > \sum_{i\in \bar{\mathcal{S}}}\|\mathbf{v}_i\|_2
\end{equation}
for any nonzero $\mathbf{v}\in \range(\mathbf{A})$, and for any partition $(\mathcal{S},\bar{\mathcal{S}})$ of $\mathcal{I}$ with $|\mathcal{S}|=s$.
\end{theorem}

\begin{IEEEproof}
See Appendix.
\end{IEEEproof}

In words, Theorem \ref{th:rsconditions} asserts that for every nonzero $\mathbf{v}\in\range(\mathbf{A})$, the sum of the $s$ smallest $\|\mathbf{v}_i\|_2$ components should be larger than the sum of the remaining $(k-s)$ components. It is worth mentioning that the range space conditions are impossible to check in practice; but they are useful in establishing identifiability, as it will be the case for the probabilistic characterization of the \eqref{eq:P0}--\eqref{eq:P1} equivalence when $\mathbf{A}$ is random (cf. Subsection \ref{subsec:prbounds}).

Another set of \eqref{eq:P0}--\eqref{eq:P1} equivalence conditions can be derived from the block restricted isometry properties of matrix $\mathbf{C}$ as defined in Remark \ref{re:MSNvsBSSR}; see \cite{ElMi09}, \cite{BaCeDuHe10}. However, these conditions are only sufficient.

\begin{remark}
Conditions \eqref{eq:rsconditions} do not depend on $\mathbf{b}$, but only on the range space of $\mathbf{A}$. Thus, whenever $\mathbf{A}$ satisfies \eqref{eq:rsconditions}, any matrix $\mathbf{A}':=\mathbf{A}\mathbf{G}$ for any nonsingular $\mathbf{G}\in \mathbb{R}^{n\times n}$ satisfies \eqref{eq:rsconditions} as well.
\end{remark}

\begin{remark}
Sufficiency of the conditions in \eqref{eq:rsconditions} remains valid even if some additional constraints of the generic form $\mathbf{x}\in \mathcal{C}$ are present in the original problem \eqref{eq:P0}. In certain applications for instance, the unknown $\mathbf{x}$ may be non-negative so that $\mathcal{C}=\mathbb{R}_{+}^n$; or, there may be a priori information of the form $\mathcal{C}=\left\{\mathbf{x}:\|\mathbf{x}-\mathbf{x}_c\|_2\leq R\right\}$, dictating the unknown vector to lie in a ball of radius $R$ around a known center $\mathbf{x}_c\in \mathbb{R}^n$. Even though the extra constraints generally reduce the feasible sets of \eqref{eq:P0} and \eqref{eq:P1}, the conditions remain sufficient. Hence, the probabilistic bound to be developed in Subsection \ref{subsec:prbounds} remains valid even when extra constraints are imposed.
\end{remark}

\subsection{Probability Bound}\label{subsec:prbounds}
As commented earlier, the conditions in Lemma \ref{le:unique-exact} are practically infeasible to check for a given sensing matrix $\mathbf{A}$. However, similar to CS \cite{CaTa05}, it will be possible to prove that the conditions in \eqref{eq:rsconditions2} hold with overwhelming probability \cite{CaTa05}, i.e., probability decaying exponentially in $n$ when $\gamma$ and $k$ are fixed, assuming $\mathbf{A}$ has i.i.d. Gaussian entries. The main result, summarized in Theorem \ref{th:bound}, is based on the following lemma.

\begin{lemma}[Deviation Inequality \cite{LeTa91}]\label{le:deviation}
Consider $\mathbf{x}\sim \mathcal{N}(\mathbf{0}_p,\mathbf{I}_p)$, and a Lipschitz continuous function $f:\mathbb{R}^p\rightarrow \mathbb{R}$ with Lipschitz constant $L$. Then for any $t\geq 0$, it holds that
\begin{equation}\label{eq:deviation}
\pr\left(f(\mathbf{x})-\mathbb{E}\left[f(\mathbf{x})\right]\leq -t\right)\leq \exp\left(-\frac{t^2}{2L^2}\right).
\end{equation}
\end{lemma}
This deviation inequality is a special case of more general concentration results \cite[Sec.~1.1]{LeTa91}. It provides exponentially decreasing bounds on the tail distribution for any sufficiently smooth function $f(\mathbf{x})$ of a multivariate Gaussian $\mathbf{x}$, thus generalizing the Chernoff bound to nonlinear functions. 

Capitalizing on Lemma \ref{le:deviation}, the next theorem extends the results of \cite[Th.~4]{StPaHa09} and its refined version \cite[Th.~3]{St09}. Focusing on the Gaussian case and following a different line of proof, neat closed-form expressions will emerge not only for the values of $\beta$ and $\gamma$, for which the probabilistic bound is valid, but also for the bound itself. The proof is based partly on the methodology of \cite{ReXuHa11}, where the minimum nuclear norm relaxation of the rank minimization problem is analyzed under linear constraints on the unknown matrix. In contrast, related probabilistic analysis in \cite{ElMi09} and \cite{BaCeDuHe10} is based on a generalization of the restricted isometry property of $\mathbf{A}$ that serves only as a sufficient condition for the exactness of the convex relaxation; see also \cite{CaTa05}.

\begin{theorem}\label{th:bound}
Let vector $\mathbf{x}_0$ be a minimizer of \eqref{eq:P0} satisfying $s>k/2$ out of the $k$ subsystems, and assume that the entries of $\mathbf{A}\in \mathbb{R}^{km\times n}$ are independently drawn from $\mathcal{N}\left(0,1\right)$.  If
\begin{equation}\label{eq:dimconditions}
\beta> \frac{\sqrt{\gamma}+1}{2}
\end{equation}
then whenever $m\geq \frac{\beta \log(e/\beta)}{(1-\alpha)c_0(\beta,\gamma)\gamma}$, the vector $\mathbf{x}_0$ is the unique minimizer of \eqref{eq:P1} with probability exceeding $1-e^{-\alpha c_0(\beta,\gamma)n+o_n(n)}$, where $c_0(\beta,\gamma):=\frac{1}{2}\left(\frac{2\beta-1}{\sqrt{\gamma}}-1\right)^2$ and $\alpha\in (0,1)$.
\end{theorem}

\begin{IEEEproof}
To lower bound the probability of success for the \eqref{eq:P1} problem, it suffices to upper bound the probability that the conditions in \eqref{eq:rsconditions} fail, an event denoted by $\mathcal{E}$. Let $\{\mathcal{S}_j\}$ be all the $N:=\binom{k}{s}$ subsets of $\mathcal{I}$ having cardinality $s$. Moreover, let $\mathcal{E}_j$ denote the event of having the conditions in \eqref{eq:rsconditions} failing for the partition $(\mathcal{S}_j,\bar{\mathcal{S}}_j)$
\begin{equation}\label{eq:Ej}
\mathcal{E}_j:=\left\{\exists \mathbf{v}\in \range(\mathbf{A})\setminus \{0\} ~ \textrm{such that} ~ \sum_{i\in\mathcal{S}_j}\|\mathbf{v}_i\|_2 \leq \sum_{i\in \bar{\mathcal{S}}_j}\|\mathbf{v}_i\|_2\right\}
\end{equation}
for $j=1,\ldots,N$. The probability of failure can be expressed as $\pr\left(\mathcal{E}\right) = \pr\left(\bigcup_{j}\mathcal{E}_j\right)$. The events $\{\mathcal{E}_j\}_{j=1}^N$ are not independent, but $\pr (\mathcal{E})$ can be bounded as
\begin{equation}\label{eq:unionbound}
\pr\left(\mathcal{E}\right)\overset{(a)}{\leq} \sum_{j=1}^{N}\pr\left(\mathcal{E}_j\right) \overset{(b)}{=} \dbinom{k}{s}\pr\left(\mathcal{E}_j\right)\overset{(c)}{\leq} e^{s(1-\log \beta)}\pr\left(\mathcal{E}_j\right)
\end{equation}
where inequality $(a)$ comes from the union bound; $(b)$ is due to the symmetry of the distribution of $\mathbf{A}$ which implies that all the $\mathcal{E}_j$'s are equiprobable; and $(c)$ is the standard upper bound of the binomial coefficient $\dbinom{k}{s}\leq \left(\dfrac{ke}{s}\right)^s$. Based on \eqref{eq:unionbound}, the goal now is to upper bound the probability $\pr (\mathcal{E}_j)$. For notational simplicity, the partition corresponding to $\mathcal{E}_j$ will be denoted by $(\mathcal{S},\bar{\mathcal{S}})$ instead of $(\mathcal{S}_j,\bar{\mathcal{S}}_j)$.

Given that $\mathbf{v}\in\range(\mathbf{A})\setminus \{0\}$, there exists a nonzero $\mathbf{u}\in\mathbb{R}^n$ such that $\mathbf{v}_i=\mathbf{A}_i\mathbf{u}$ for $i=1,\ldots,k$. To render the inequality in \eqref{eq:Ej} scale-invariant, one can study only the cases for which $\|\mathbf{u}\|_2=1$; hence,
\begin{align}
\pr\left(\mathcal{E}_j\right)&=\pr\left(\exists \mathbf{u} ~\textrm{with}~ \|\mathbf{u}\|_2=1 ~ \textrm{such that} ~ \sum_{i\in\mathcal{S}}\|\mathbf{A}_i\mathbf{u}\|_2 - \sum_{i\in \bar{\mathcal{S}}}\|\mathbf{A}_i\mathbf{u}\|_2 \leq 0\right)\label{eq:Es2extra}\\
&= \pr\bigg(f(\mathbf{A})\leq 0\bigg)\label{eq:Es2}
\end{align}
where 
\begin{equation}\label{eq:fA}
f\left(\mathbf{A}\right):=\inf_{\|\mathbf{u}\|_2=1} \left\{ \sum_{i\in \mathcal{S}}\|\mathbf{A}_i\mathbf{u}\|_2 - \sum_{i\in \bar{\mathcal{S}}}\|\mathbf{A}_i\mathbf{u}\|_2\right\}.
\end{equation}
The equality from \eqref{eq:Es2extra} to \eqref{eq:Es2} comes from the fact that if there exists a unit $\ell_2$-norm $\mathbf{u}$ satisfying the inequality in \eqref{eq:Es2extra}, then the minimizer of $f\left(\mathbf{A}\right)$ should also satisfy this property. The function $f(\mathbf{A})$ possesses convenient properties which facilitate the application of Lemma \ref{le:deviation}. Specifically, it is shown in the Appendix that: $f(\mathbf{A})$ is Lipschitz continuous with constant $L\leq \sqrt{k}$ (cf. Lemma \ref{le:lipschitz}); and the expected value of the function is lower bounded (cf. Lemma \ref{le:meanlb}), that is $\mathbb{E}\left[f(\mathbf{A})\right]\geq \mu =  \left(\frac{2\beta-1}{\sqrt{\gamma}}-1\right) \sqrt{kn}\left(1+o_n(1)\right)$. Hence, for every $t\geq 0$, Lemma \ref{le:deviation} implies that
\begin{equation}\label{eq:prbound}
\pr\bigg(f(\mathbf{A})\leq \mu-t\bigg) \leq
\pr\bigg(f(\mathbf{A})\leq \mathbb{E}\left[f(\mathbf{A})\right]-t\bigg)\leq.
e^{-t^2/(2k)}
\end{equation}
Upon focusing on $\mu$ and ignoring the $o_n(1)$ term, whenever $\beta>(\sqrt{\gamma}+1)/2$ so that $\mu> 0$, and setting $t=\mu$ in \eqref{eq:prbound}, yields the bound
\begin{equation}\label{eq:weakbound}
\pr(\mathcal{E}_j)\leq \pr\bigg(f(\mathbf{A})\leq 0\bigg) \leq e^{-c_0(\beta,\gamma) n + o_n(n)}
\end{equation}
where $c_0(\beta,\gamma):=\left(\frac{2\beta-1}{\sqrt{\gamma}}-1\right)^2/2$.

Substituting the bound \eqref{eq:weakbound} into \eqref{eq:unionbound}, it follows that
\begin{align}
\pr(\mathcal{E})&\leq \exp\left(-\left(c_0(\beta,\gamma)-\frac{s\log(e/\beta)}{n}\right)n + o_n(n)\right)\nonumber\\
&\leq \exp\left(-c_1(\beta,\gamma)n +o_n(n)\right).\label{eq:bound}
\end{align}
For every $\beta>\left(\sqrt{\gamma}+1\right)/2$, choose $c_1(\beta,\gamma)=\alpha c_0(\beta,\gamma)$, and define $c_2(\beta,\gamma):=\left((1-\alpha)c_0(\beta,\gamma)\right)^{-1}$ for any $\alpha\in(0,1)$. Then, whenever $m\geq c_2(\beta,\gamma) \beta \log(e/\beta)/\gamma$, the bound in \eqref{eq:bound} is nontrivial.
\end{IEEEproof}

%The following corollary is an immediate consequence of the Borel-Cantelli lemma.
%
%\begin{corollary}[Almost sure convergence]\label{co:as_convergence}
%Whenever the quadruplet $(n,m,k,s)$ satisfies the conditions of Theorem \ref{th:bound}, the solution of \eqref{eq:P1} coincides with the solution of \eqref{eq:P0} for a random matrix $\mathbf{A}$ drawn from the Gaussian ensemble almost surely as $n\rightarrow \infty$.
%\end{corollary}

\begin{remark}
As a sanity test, the condition $\beta>(\sqrt{\gamma}+1)/2$ posed by Theorem \ref{th:bound} coincides with that in \cite[Th.~3]{St09} after the appropriate mapping of dimensions. However, in Theorem \ref{th:bound}, both the values of $m$ over which the bound holds, as well as the bound itself are explicitly defined.
\end{remark}

\begin{remark}
As expected, the condition $\beta>(\sqrt{\gamma}+1)/2$ is clearly stronger than the condition $\beta>(\gamma+1)/2$ implied by the uniqueness of the \eqref{eq:P0} solution in Lemma \ref{le:unique}.
\end{remark}

\section{Robustness to Noise}\label{sec:RSN}
In a more realistic sensing scenario, the acquired measurements are corrupted by additive noise. If $\mathcal{S}_0$ denotes the unknown subset of reliable sensors, the pertinent model is 
\begin{equation}\label{eq:noisy_model}
\mathbf{b}_{\mathcal{S}_0}=\mathbf{A}_{\mathcal{S}_0}\mathbf{x}_0+\mathbf{n}_{\mathcal{S}_0}
\end{equation}
where $\mathbf{n}_{\mathcal{S}_0}$ stands for zero-mean noise assumed independent across sensors. Vector $\mathbf{n}_{\mathcal{S}_0}$ models ambient noise, finite precision, analog-to-digital conversion, and quantization effects, communication noise, or even, the inadequacy of linear regression to fully capture the measured data $\mathbf{b}_{\mathcal{S}_0}$.

In this noisy case, the unknown $\mathbf{x}_0$ does not exactly satisfy the linear subsystems in $\mathcal{S}_0$. In an attempt to exploit the link between \eqref{eq:P1} and \eqref{eq:BBP} when noise is present, one may be tempted to apply the group-Lasso regularization, which was originally proposed for recovering block-sparse vectors in a linear regression setup \cite{YuLi06}. However, this approach is not applicable because $\mathbf{r}$ is not block sparse when noise is present. In fact, solvers of the noise-free setups \eqref{eq:P1} and \eqref{eq:P2} are useful for analyzing uniqueness and identifiability issues. In addition, \eqref{eq:P1} and \eqref{eq:P2} solvers are practically suitable for high-SNR sensing applications. This motivates the ensuing framework which is suitable for RS in the presence of noise. Without additional prior information on the model describing the unreliable sensors, the noisy counterpart of the RS problem can be stated as follows.

\begin{definition}[Robust sensing in noise (RSN)]\label{def:RSN}
Given $\left\{\mathbf{b}_i, \mathbf{A}_i \right\}_{i\in\mathcal{I}}$ where $\mathbf{b}_i\in\mathbb{R}^{m}$ and $\mathbf{A}_i\in\mathbb{R}^{m\times n}$, for which an unknown subset $\mathcal{S}_0\subset \mathcal{I}$ of known cardinality $s$ follows the model in \eqref{eq:noisy_model}, estimate the unknown $\mathbf{x}_0$ by minimizing the least-squares error over any $\mathcal{S}\subset \mathcal{I}$ with $|\mathcal{S}|=s$.
\end{definition}

The aforementioned problem statement lends itself naturally to the following optimization problem
\begin{equation}\label{eq:RSN1}
\min_{\mathbf{x}}\min_{|\mathcal{S}|=s} \|\mathbf{b}_{\mathcal{S}}-\mathbf{A}_{\mathcal{S}}\mathbf{x}\|_{2}^2.
\end{equation}
The function of $\mathbf{x}$ defined by the inner minimization is the pointwise minimum over finitely many convex functions, and as such, it is non-convex. Solving \eqref{eq:RSN1} incurs combinatorial complexity since one has to solve all the $\binom{k}{s}$ LS problems before solving the outer minimization.

An optimization problem related to that in \eqref{eq:RSN1} is the following
\begin{subequations}\label{eq:RSN2}
\begin{align}
\min_{\mathbf{x}}~&~\sum_{i=1}^k h(\mathbf{b}_i-\mathbf{A}_i\mathbf{x})\label{eq:RSN2a}\\
\textrm{s.t.}~&~h(\mathbf{r}_i):=\left\{\begin{array}{ll}
\frac{1}{2}\|\mathbf{r}_i\|_2^2 &,~\|\mathbf{r}_i\|_2\leq \lambda\\
\frac{1}{2}\lambda^2&,~\|\mathbf{r}_i\|_2> \lambda
\end{array}\right.,~\lambda\geq 0.\label{eq:RSN2b}
\end{align}
\end{subequations}
Functional $h(\mathbf{r}_i)$ amounts to the LS cost for residuals smaller than the threshold $\lambda$, and ignores sensors attaining larger residuals. In the scalar case (cf. $m=1$), problem \eqref{eq:RSN2} has been considered in \cite[Sec.~6.1.2]{BoVa04}. Problems \eqref{eq:RSN1} and \eqref{eq:RSN2} are related as follows: suppose that for a specific $\lambda$ the solution of \eqref{eq:RSN2} is $\mathbf{x}^{\star}$ for which there are $s^{\star}$ residuals satisfying the upper branch of \eqref{eq:RSN2b}. Then it can be readily shown that $\mathbf{x}^{\star}$ is a solution of \eqref{eq:RSN1} for $s=s^{\star}$. Unfortunately though, $h(\mathbf{r}_i)$ is non-convex as well. The problem in \eqref{eq:RSN2} can be surrogated by replacing $h(\mathbf{r}_i)$ by its closest convex approximation, which is pursued in the next subsection by establishing a neat link between the RSN problem at hand and robust estimation methods \cite[Ch.~7]{HuRo09}, \cite[Ch.~4]{MaMaYo06}.

\subsection{RSN and Robust Linear Multivariate Regression}\label{subsec:solving_RSN}
Building on Remark \ref{re:MSNvsL1} of Subsection \ref{subsec:RS_convex}, the unreliable sensors can be viewed as giving rise to outlier-corrupted equations in a linear regression setting. Robust linear regression has been extensively studied over the past decades \cite{HuRo09}, \cite{MaMaYo06}. 

When $m=1$, the RSN problem can be solved by Huber's M-estimator
\begin{subequations}\label{eq:huber_problem}
\begin{align}
\hat{\mathbf{x}}=\arg\min_{\mathbf{x}} ~&~\sum_{i=1}^k \rho(b_i-\mathbf{a}_i^T\mathbf{x})\label{eq:huber}\\
\textrm{s.t.}~&~\rho(r):=\left\{\begin{array}{ll}
\frac{1}{2}r^2 &,~|r|\leq \tau\\
\tau |r|-\frac{\tau^2}{2} &,~|r|>\tau
\end{array}\right.\label{eq:huber_function}
\end{align}
\end{subequations}
where $\rho(r)$ is the Huber function for $\tau>0$. The problem in \eqref{eq:huber_problem} is convex, and can be cast as an SOCP \cite{Fu99}, \cite{MaMu00}, \cite[p.~190]{BoVa04}. Regarding the cutoff parameter $\tau$, when the outliers' distribution is known a priori, its value is available in closed form so that Huber's M-estimator is asymptotically optimal; see  \cite[Sec.~4.5]{HuRo09}. Alternatively, assuming that the noise is standard Gaussian, $\tau$ is usually set to $\tau=1.34$ such that the estimator in \eqref{eq:huber_problem} is 95\% asymptotically efficient at the normal distribution \cite[p.~26]{MaMaYo06}. To render Huber's M-estimator invariant to any noise variance $\sigma^2$, one has to multiply $\tau$ by $\sigma$ in \eqref{eq:huber_function}. If $\sigma$ is unknown, a robust estimate of it is commonly used instead \cite[Sec.~4.4]{MaMaYo06}. 

The case $m>1$, which is of interest here falls under the realm of robust multivariate linear regression \cite{AeWi05}, \cite{BaRaWu92}. The novel approach to tackle it will be to postulate a model accommodating inconsistent sensors, approximate the meaningful cost of \eqref{eq:RSN2} by a convex one, and solve it using an efficient globally convergent algorithm.

Consider modeling the unreliable sensors using the auxiliary outlier vectors $\{\mathbf{u}_i\in \mathbb{R}^m\}_{i=1}^k$. Vector $\mathbf{u}_i=\mathbf{0}$ if the $i$-th sensor is reliable; and $\mathbf{u}_i\neq\mathbf{0}$ deterministically, otherwise. Model \eqref{eq:noisy_model} can now be extended to incorporate the unreliable sensors as
\begin{equation}\label{eq:noisy_outlier_model}
\mathbf{b}_i=\mathbf{A}_i\mathbf{x}+\mathbf{u}_i + \mathbf{n}_i,~i=1,\ldots,k.
\end{equation}
Since some $\mathbf{u}_i$'s are zero, the aggregate outlier vector $\mathbf{u}^T:=[\mathbf{u}_1^T~\cdots~\mathbf{u}_k^T]$ is block sparse. Hence, using the aggregate model $\mathbf{b}= \mathbf{A}\mathbf{x} + \mathbf{u}+\mathbf{n}$, the novel RSN solver amounts to \begin{empheq}[box=\fbox]{align}\tag{$P_3$}\label{eq:P3}
\min_{\mathbf{x},\mathbf{u}} \frac{1}{2} \|\mathbf{b}- \mathbf{A}\mathbf{x}-\mathbf{u}\|_2^2 + \lambda\sum_{i=1}^k \|\mathbf{u}_i\|_2
\end{empheq}
where $\lambda>0$ is an appropriately chosen tuning parameter. Among the two optimization variables of \eqref{eq:P3}, only the outlier vector $\mathbf{u}$ is block sparse. For $m=1$, \eqref{eq:P3} reduces to the cost proposed in \cite{Fu99} and shown to be equivalent to \eqref{eq:huber_problem}. Even when the initial matrix of interest $\mathbf{A}$ is tall, \eqref{eq:P3} always entails the fat matrix $[\mathbf{A}~\mathbf{I}_{km}]\in \mathbb{R}^{km\times (n+km)}$. The second part is a regularization term, reminiscent of the group Lasso penalty function \cite{YuLi06}, which is known to promote block sparsity in the $\mathbf{u}$ vector. The latter will be explicitly accounted for in the forthcoming analysis.

\subsection{Solving \eqref{eq:P3}}
To better understand \eqref{eq:P3} and develop an efficient solver, it is prudent to explore the form of its minimizer(s). Let $[(\mathbf{x}^{\star})^T~ (\mathbf{u}^{\star})^T]^T$ denote a minimizer of \eqref{eq:P3}, and define the associated residual vector $\mathbf{r}^{\star}:=\mathbf{b}- \mathbf{A}\mathbf{x}^{\star}$. Given $\mathbf{x}^{\star}$, the vectors $\{\mathbf{u}^{\star}_i\}_{i=1}^k$ in \eqref{eq:P3} can be found separately as the minimizers of
\begin{align}
\min_{\mathbf{u}_i} ~&~ \phi(\mathbf{u}_i)\label{eq:ui}\\
\textrm{s.t.} ~&~ \phi(\mathbf{u}_i):=\frac{1}{2} \|\mathbf{r}_i^{\star}-\mathbf{u}_i\|_2^2 + \lambda\|\mathbf{u}_i\|_2,~i=1,\ldots,k.\nonumber
\end{align}
Although $\phi(\mathbf{u}_i)$ is not everywhere differentiable, its subdifferential $\partial \phi(\mathbf{u}_i)$ can be defined \cite{BoVa04}. For $\mathbf{u}_i\neq \mathbf{0}$, where $\phi(\mathbf{u}_i)$ is differentiable, the subdifferential is simply $\mathbf{u}_i\left(1+\lambda/\|\mathbf{u}_i\|_2\right)-\mathbf{r}_i^{\star}$. Otherwise, by definition and after using \eqref{eq:norm}, $\partial \phi(\mathbf{u}_i)$ can be shown to be the set $\left\{\lambda\mathbf{g}_i-\mathbf{r}^{\star}_i\right\}$ $\forall$ $\|\mathbf{g}_i\|_2\leq 1$. Compactly, 
\begin{equation}\label{eq:subdiff}
\partial \phi(\mathbf{u}_i):=\left\{\begin{array}{ll}
\mathbf{u}_i\left(1+\frac{\lambda}{\|\mathbf{u}_i\|_2}\right)-\mathbf{r}^{\star}_i&,~\mathbf{u}_i\neq \mathbf{0}\\
\left\{\lambda\mathbf{g}_i-\mathbf{r}^{\star}_i:~\|\mathbf{g}_i\|_2\leq 1\right\} &,~\mathbf{u}_i=\mathbf{0}.
\end{array}\right.
\end{equation}
Vector $\mathbf{u}^{\star}_i$ is a minimizer of \eqref{eq:ui} if and only if $\mathbf{0}\in \partial \phi(\mathbf{u}^{\star}_i)$. Based on \eqref{eq:subdiff}, two cases are considered.

First, if $\mathbf{u}_i^{\star}\neq\mathbf{0}$, the condition $\mathbf{0}\in \partial \phi(\mathbf{u}^{\star}_i)$ yields 
\begin{equation}\label{eq:uidummy}
\mathbf{u}^{\star}_i(1+\lambda/\|\mathbf{u}_i^{\star}\|_2) =\mathbf{r}^{\star}_i
\end{equation}
which means that $\mathbf{u}_i^{\star}$ is a positively scaled version of $\mathbf{r}_i^{\star}$. Considering the $\ell_2$-norm in both sides of \eqref{eq:uidummy}, it follows that $\|\mathbf{u}_i^{\star}\|_2=\|\mathbf{r}_i^{\star}\|_2-\lambda$. Plugging $\|\mathbf{u}_i^{\star}\|_2$ back into \eqref{eq:uidummy}, yields $\mathbf{u}^{\star}_i= \mathbf{r}^{\star}_i\left(1-\frac{\lambda}{\|\mathbf{r}^{\star}_i\|_2}\right)$. Since $\|\mathbf{u}_i^{\star}\|_2>0$, this holds if and only if $\|\mathbf{r}_i^{\star}\|_2>\lambda$.

Second, for the minimizer to be $\mathbf{u}_i^{\star}=\mathbf{0}$, there should be a $\mathbf{g}_i^{\star}$ for which $\|\mathbf{g}_i^{\star}\|_2\leq 1$ and $\lambda\mathbf{g}_i^{\star}=\mathbf{r}_i^{\star}$, or equivalently, $\|\mathbf{r}_i^{\star}\|_2\leq \lambda$. The latter proves that \eqref{eq:P3} indeed admits a block-sparse minimizer $\mathbf{u}^{\star}$. 

Substituting $\mathbf{u}^{\star}_i$ into \eqref{eq:ui}, yields $\phi(\mathbf{u}^{\star}_i)= \|\mathbf{r}_i^{\star}\|_2^2/2$, when $\|\mathbf{r}^{\star}_i\|_2\leq \lambda$; and $\phi(\mathbf{u}_i^{\star})=\lambda\|\mathbf{r}_i^{\star}\|_2-\lambda^2/2$, otherwise. Having minimized \eqref{eq:P3} over the $\mathbf{u_i}$'s, the minimizer $\mathbf{x}^{\star}$ can now be found as 
\begin{subequations}\label{eq:vector_huber_problem}
\begin{align}
\min_{\mathbf{x}}~&~ \sum_{i=1}^k \rho_v(\mathbf{b}_i-\mathbf{A}_i\mathbf{x}) \label{eq:huber_like}\\
\textrm{s.t.}~&~\rho_v(\mathbf{r}_i):=\left\{\begin{array}{ll}
\frac{1}{2}\|\mathbf{r}_i\|_2^2 &,~\|\mathbf{r}_i\|_2\leq \lambda\\
\lambda \|\mathbf{r}_i\|_2-\frac{\lambda^2}{2} &,~\|\mathbf{r}_i\|_2> \lambda
\end{array}\right.\label{eq:huber_vector}
\end{align}
\end{subequations}
where $\rho_v(\mathbf{r}_i)$ is a vector-generalized Huber function. It is now evident that \eqref{eq:P3} is equivalent to \eqref{eq:vector_huber_problem}, which rather surprisingly turns out to be a generalization of Huber's M-estimator \eqref{eq:huber_problem} to the vector case. The sensors capable of achieving a lower $\|\mathbf{r}_i\|_2$ value, and are more likely to be reliable, appear in \eqref{eq:vector_huber_problem} under the conventional LS criterion. But the sensors having $\|\mathbf{r}_i\|_2>\lambda$, contribute $(\lambda\|\mathbf{r}_i\|_2-\lambda^2/2)<\|\mathbf{r}_i\|_2^2/2$ to the cost, and are deemed ``less important'' in specifying $\mathbf{x}$. For the latter set of sensors, $\mathbf{u}_i^{\star}\neq \mathbf{0}$ holds too. Thus, \eqref{eq:P3} not only estimates the unknown vector $\mathbf{x}$, but also reveals the sensors most likely to be unreliable in the presence of noise.

Regarding the cutoff parameter $\lambda$ in \eqref{eq:P3} and \eqref{eq:huber_vector}, it is worth noting that when $\lambda \rightarrow 0^+$, the costs of \eqref{eq:vector_huber_problem} and \eqref{eq:P3} tend to the cost of \eqref{eq:P1}. Consequently, for $\lambda\rightarrow 0^+$ the data of all sensors are declared to contain outliers; and according to the previous analysis, $(\mathbf{b}_i-\mathbf{A}_i\mathbf{x}^{\star})\rightarrow \mathbf{u}^{\star}_i\neq \mathbf{0}$ for all $i$. This suggests that the solution of \eqref{eq:P1} does not provide zero residuals anymore. On the other hand, as $\lambda \rightarrow \infty$, the same costs reduce to the LS criterion, and all sensors are classified as reliable, or $\mathbf{u}_i^{\star}=\mathbf{0}$ for all $i$.

A heuristic rule of thumb for practically selecting $\lambda$ is setting it to $\tau\sqrt{m}$, where $\tau$ is the equivalent parameter for the scalar case and has been selected according to the techniques mentioned after \eqref{eq:huber_problem}. If the number of reliable sensors is roughly known (e.g., based on prior operation of the network), an alternative approach is solving \eqref{eq:P3} for a grid of $\lambda$ values and selecting the one identifying the prescribed number of outliers. Note that solving \eqref{eq:P3} for several values of $\lambda$ can be efficiently performed either through the group-LARS algorithm \cite{YuLi06}, or, by using the block coordinate descent algorithm of the next subsection with what is called ``warm startup'' \cite{FrHaTi10}. The latter initializes the tentative solutions of \eqref{eq:P3} for a grid value of $\lambda$ with the solution derived for the previous grid value of $\lambda$. The computational efficiency of such an approach has been numerically verified for the Lasso problem \cite{FrHaTi10}, \cite{Ti96}.

\begin{remark}\label{re:colored}
In Problem Statement \ref{def:RSN}, the noise term was assumed to be independent across sensors. Specifications such as the geographical distribution of sensors may impose correlation across different sensor readings. In this case, if the covariance matrix $\mathbf{\Sigma}$ of the aggregate noise vector $\mathbf{n}^T: =[\mathbf{n}_1^T~\cdots~\mathbf{n}_k^T]$ is known, a standard preprocessing step is to prewhiten the data as $\mathbf{b}':=\mathbf{\Sigma}^{-1/2}\mathbf{b}$ and $\mathbf{A}':=\mathbf{\Sigma}^{-1/2}\mathbf{A}$. Prewhitening ``spreads'' the influence of unreliable sensors across the entries of $\mathbf{b}'$. As a result, the LS and $\ell_1$-error regression estimators and even the robust Huber M-estimator are not applicable; see also \cite{Fu99} for similar observations in the scalar case ($m=1$). On the contrary, given that $\mathbf{u}$ remains block sparse, the \eqref{eq:P3} estimator can successfully handle a colored noise setup by simply modifying its cost to $\|\mathbf{\mathbf{b}' -\mathbf{A}'\mathbf{x}- \Sigma}^{-1/2}\mathbf{u}\|_2^2/2 + \lambda\sum_{i=1}^k \|\mathbf{u}_i\|_2$.
\end{remark}

\subsection{A Block Coordinate Descent Algorithm}\label{subsec:bcd_algorithm}
As mentioned earlier, \eqref{eq:P3} is convex. It can be cast as an SOCP and solved by standard, interior point-based solvers. An alternative solver of \eqref{eq:P3} exploiting the problem structure and offering computational advantages is block coordinate descent, which has been successfully applied to related optimization problems \cite{FrHaHoTi07}, \cite{WuLa08}. The core idea behind this solver is to partition the optimization variable into blocks, and minimize iteratively the cost w.r.t. one block variable while keeping the rest fixed. 

To apply block coordinate descent to the RSN problem at hand, consider minimizing the cost separately w.r.t. $\mathbf{x}$ and $\mathbf{u}$. Each iteration involves two steps: In the first step, the objective is minimized w.r.t. $\mathbf{x}$, while keeping $\mathbf{u}$ fixed, whereas in the second step the roles are interchanged. Specifically, let $\mathbf{x}^{(l-1)}$ and $\mathbf{u}^{(l-1)}$ denote the tentative solutions at the $(l-1)$-th iteration. During the first step of the $l$-th iteration, fix $\mathbf{u}=\mathbf{u}^{(l-1)}$, and find $\mathbf{x}^{(l)}$ as the minimizer of the resultant quadratic; that is,
\begin{equation}\label{eq:x_solution}
\mathbf{x}^{(l)}=(\mathbf{A}^T\mathbf{A})^{-1}\mathbf{A}^T(\mathbf{b}-\mathbf{u}^{(l-1)}).
\end{equation} 
In the second step, fix $\mathbf{x}{=}\mathbf{x}^{(l)}$ and find the $\mathbf{u}_i^{(l)}$'s as the minimizers of the per-sensor optimization problems
\begin{equation}\label{eq:lresiduals}
\min_{\mathbf{u}_i}~\frac{1}{2}\|\mathbf{r}_i^{(l)}-\mathbf{u}_i\|_2^2+\lambda\|\mathbf{u}_i\|_2
\end{equation}
where $\mathbf{r}_i^{(l)}:= \mathbf{b}_i-\mathbf{A}_i\mathbf{x}^{(l)}$ for $i=1,\ldots,k$. As per \eqref{eq:ui}, the solutions of \eqref{eq:lresiduals} are provided neatly in closed form\footnote{This is not the case for the colored noise scenario discussed in Remark \ref{re:colored}, where the vectors $\{\mathbf{u}_i\}$ can then be jointly found by any group Lasso algorithm instead \cite{YuLi06}.} as
\begin{equation}\label{eq:u_solution}
\mathbf{u}_i^{(l)}=\left\{\begin{array}{ll}
\mathbf{0} &,~\|\mathbf{r}_i^{(l)}\|_2\leq \lambda\\
\mathbf{r}_i^{(l)}\left(1-\frac{\lambda}{\|\mathbf{r}_i^{(l)}\|_2}\right) &,~\|\mathbf{r}_i^{(l)}\|_2> \lambda.
\end{array}\right.
\end{equation}
The solution in \eqref{eq:u_solution} does not require $\mathbf{x}^{(l)}$, but only $\mathbf{r}^{(l)}$. Combining \eqref{eq:x_solution} and \eqref{eq:lresiduals}, it follows that
\begin{equation}\label{eq:lresiduals2}
\mathbf{r}^{(l)}=\mathbf{P}_{A}^{\bot}\mathbf{b} + \mathbf{P}_{A}\mathbf{u}^{(l-1)}
\end{equation}
where $\mathbf{P}_{A} := \mathbf{A} (\mathbf{A}^T\mathbf{A})^{-1} \mathbf{A}^T$ and $\mathbf{P}^{\bot}_{A} :=\mathbf{I} - \mathbf{P}_{A}$.

Summarizing, the iterations entail: (a) updating the residuals based on \eqref{eq:lresiduals2}; and (b) applying the thresholding rule in \eqref{eq:u_solution}. As matrix $\mathbf{P}_{A}$ and vector $\mathbf{P}_{A}^{\bot} \mathbf{b}$ can be computed offline, the most computationally demanding operation is the matrix-vector product in step (a). Since $km\geq n$, this product would better be implemented as $\left(\mathbf{A} (\mathbf{A}\mathbf{A}^T)^{-1} \right)\left(\mathbf{A}^T\mathbf{u}\right)$ in $O(kmn)$ operations. The developed algorithm has overall complexity $O(kmn)$ per iteration. The presence of zero blocks in $\mathbf{u}$ can be further exploited to save computations. Numerical simulations demonstrate that the overall complexity of this block-coordinate approach is much lower than the complexity of the interior point-based algorithms.

Due to the specific form of \eqref{eq:P3}, convergence of the block coordinate descent iteration follows readily from the results of \cite{Ts01}. The algorithm can be initialized at $\mathbf{u}^{(0)} = \mathbf{0}$, so that $\mathbf{x}^{(1)}$ is the conventional LS solution. It is terminated when the relative error $\|\mathbf{u}^{(l)} - \mathbf{u}^{(l-1)}\|_2/\|\mathbf{u}^{(l)}\|_2$ becomes smaller than a predefined threshold, e.g., $\epsilon=10^{-6}$. Upon termination, the output is the solution vector $\hat{\mathbf{u}}$, which reveals the sensors affected by outliers, whereas the solution $\hat{\mathbf{x}}$ can be obtained directly from \eqref{eq:x_solution}.

\subsection{A Non-Convex Surrogate for RSN}\label{subsec:non_convex}
In the context of robust linear regression, Huber's M-estimator is just one choice from the class of robust estimators defined as the minimizers of \eqref{eq:huber_problem} for appropriately chosen $\rho$ functions. It has been argued that estimators corresponding to non-convex $\rho$ functions, such as the bisquare (Tukey's), Hampel's, or Andrew's estimators, yield improved robustness-efficiency trade-offs in practice \cite[p.~99]{MaMaYo06}. Similarly in the multivariate case, convex M-estimators \cite{BaRaWu92} are practically replaced by non-convex M- or S-estimators appropriately initialized \cite{AeWi05}.

Alternatively, it is of interest to explore a non-convex surrogate of \eqref{eq:P3} paralleling that of Subsection \ref{subsec:RS_concave}. Recall that the RSN solver in \eqref{eq:P3} seeks $\mathbf{x}$ and $\mathbf{u}$ based on fewer observations than unknowns, but taking advantage of $\mathbf{u}$'s block sparsity. To further promote block sparsity in $\mathbf{u}$, the $\|\mathbf{u}_i\|_2$ terms in \eqref{eq:P3} can be replaced by $\log(\|\mathbf{u}_i\|_2+\delta)$ for a small positive $\delta$, to end up with the non-convex problem
\begin{empheq}[box=\fbox]{align}\tag{$P_4$}\label{eq:P4}
\min_{\mathbf{x},\mathbf{u}} \frac{1}{2} \|\mathbf{b}- \mathbf{A}\mathbf{x}-\mathbf{u}\|_2^2 + \lambda\sum_{i=1}^k \log(\|\mathbf{u}_i\|_2+\delta).
\end{empheq}
Following the majorization-minimization rationale presented in Subsection \ref{subsec:RS_concave}, \eqref{eq:P4} can be driven to a stationary point \cite{LaHuYa00} using the iterations
\begin{align}
\left(\mathbf{x}^{(l)},\mathbf{u}^{(l)}\right):=
\arg &\min_{\mathbf{x},\mathbf{u}}
\frac{1}{2}\|\mathbf{b}-\mathbf{A}\mathbf{x}-\mathbf{u}\|_2^2 + \lambda\sum_{i=1}^k w_i^{(l)}\|\mathbf{u}_i\|_2,\label{eq:P4a}\\
w_i^{(l)}&:=\left(\|\mathbf{u}_i^{(l-1)}\|_2+\delta\right)^{-1},~i=1,\ldots,k.\nonumber
\end{align}

The optimization per iteration of \eqref{eq:P4a} is a weighted version of \eqref{eq:P3}, and thus can be efficiently solved using the steps \eqref{eq:x_solution} and \eqref{eq:u_solution} after replacing $\lambda$ in \eqref{eq:u_solution} with $\lambda_i^{(l)}:=\lambda w_i^{(l)}$ for all $i$ at the $l$-th iteration. The iterations can be initialized with the \eqref{eq:P3} solution which corresponds to setting all weights to unity. The simulations of Section \ref{sec:simulations} will demonstrate that the \eqref{eq:P4} solver outperforms that of \eqref{eq:P3} in terms of the mean-square error (MSE) even after a single iteration. Note that as with \eqref{eq:P4a}, single-iteration methods based on non-convex surrogates of the (group) Lasso cost function have been proposed with well documented properties \cite{ZouLi08}, \cite{NaRi08}.

\section{Simulated Tests}\label{sec:simulations}
\subsection{Checking the Weak Bound}\label{subsec:sim-weak}
Among the results of Section \ref{sec:performance}, the one that can be numerically validated is the weak bound of \eqref{eq:weakbound}. This bound is termed weak because it refers to the occurrence of a single event $\mathcal{E}_s$, namely, to a single partition $(\mathcal{S},\bar{\mathcal{S}})$ with $\mathcal{S}=s$. According to this bound, if $\beta$ and $\gamma$ are kept fixed and as long as $\beta>(\sqrt{\gamma}+1)/2$, the probability $\pr(\mathcal{E}_s)$ is arbitrarily small for large $n$.

%%% \label{subfig:weakbound4}}

To validate this result, the entries of $\mathbf{A}$ are drawn independently from $\mathcal{N}(0,1)$, and the unknown vector is modeled as $\mathbf{x}_0\sim n^{-1/2}\mathcal{N}(\mathbf{0},\mathbf{I}_n)$. Given that $\pr(\mathcal{E}_s)$ is invariant to the permutations of the subsystems, the partition $(\mathcal{S}_0,\bar{\mathcal{S}}_0)$ with $\mathcal{S}_0=\{1,\ldots,s\}$ is simply selected. The output of the consistent subsystems is $\mathbf{b}_{\mathcal{S}_0}=\mathbf{A}_{\mathcal{S}_0}\mathbf{x}_0$; whereas for the inconsistent ones $\mathbf{b}_{\bar{\mathcal{S}}_0}=\mathbf{w}$ is simulated with $\mathbf{w}\sim \mathcal{N}(\mathbf{0},\mathbf{I}_{(k-s)m})$. Notice that due to the selected normalization, the observation vectors have equal variance, i.e., $\mathbb{E}[\|\mathbf{b}_i\|_2^2]=m$ for all $i\in \mathcal{I}$. For several $(n,m)$ pairs, ten values of $\gamma$ are selected uniformly over the interval $\left(0.1,1\right]$ that correspond to ten values of $k$. And for every $\gamma(k)$, the number of consistent subsystems $s$ is chosen such that $\beta(s,k)=s/k\in\left[0.5,1\right]$. For each pair $(\gamma(k),\beta(s,k))$, the probability of \eqref{eq:P1} identifying uniquely the \eqref{eq:P0} solution is empirically evaluated through 100 Monte Carlo runs. For each experiment, the solution of \eqref{eq:P1} is deemed successful whenever $\hat{\mathbf{x}}$ satisfies $\|\hat{\mathbf{x}}-\mathbf{x}_0\|_{\infty}\leq 10^{-4}$.

\begin{figure}[!t]
\centering
\subfigure[$n=40,~m=20$.]{
\includegraphics[width=0.4\linewidth]{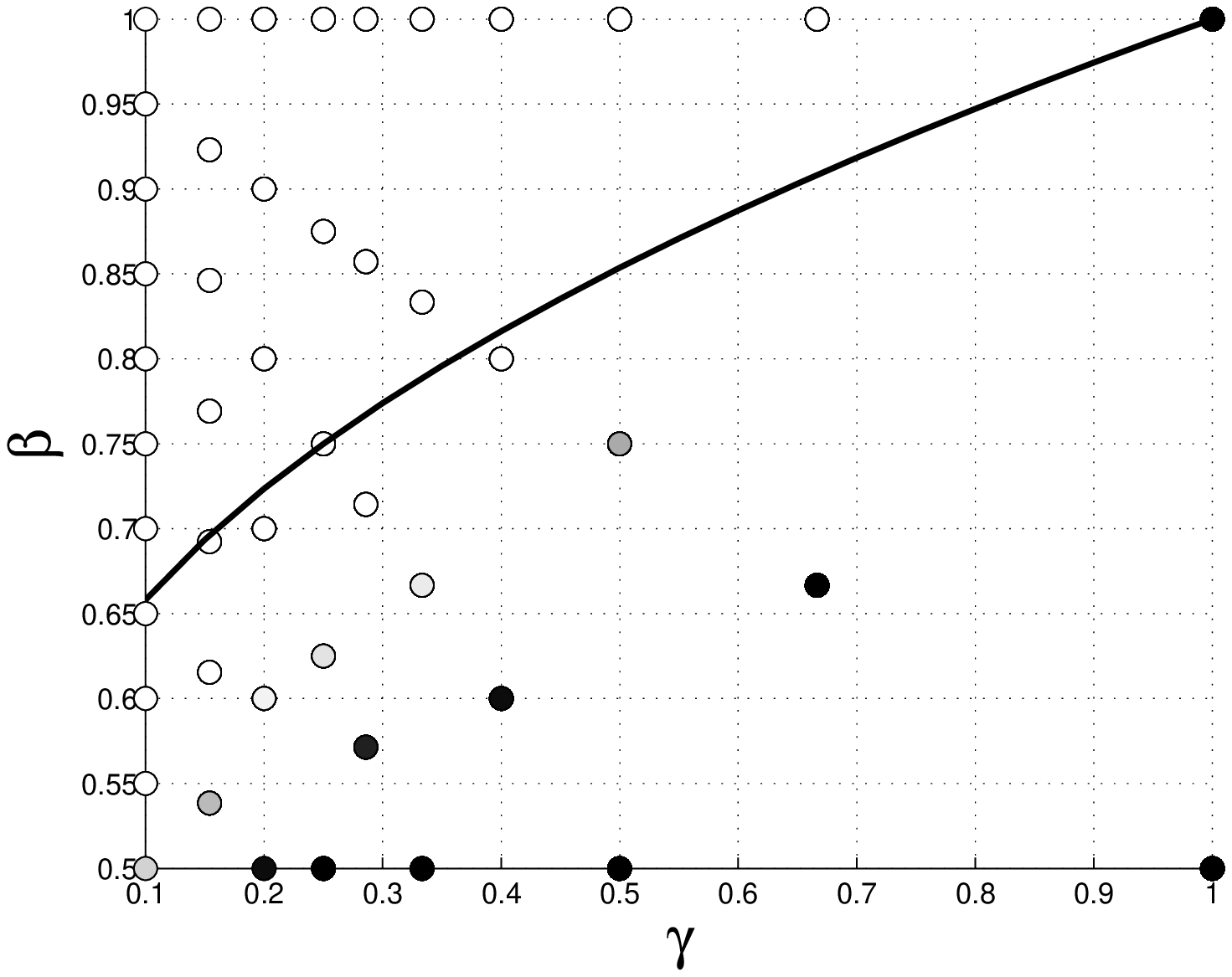}
\label{subfig:weakbound1}}
\subfigure[$n=40,~m=10$.]{
\includegraphics[width=0.4\linewidth]{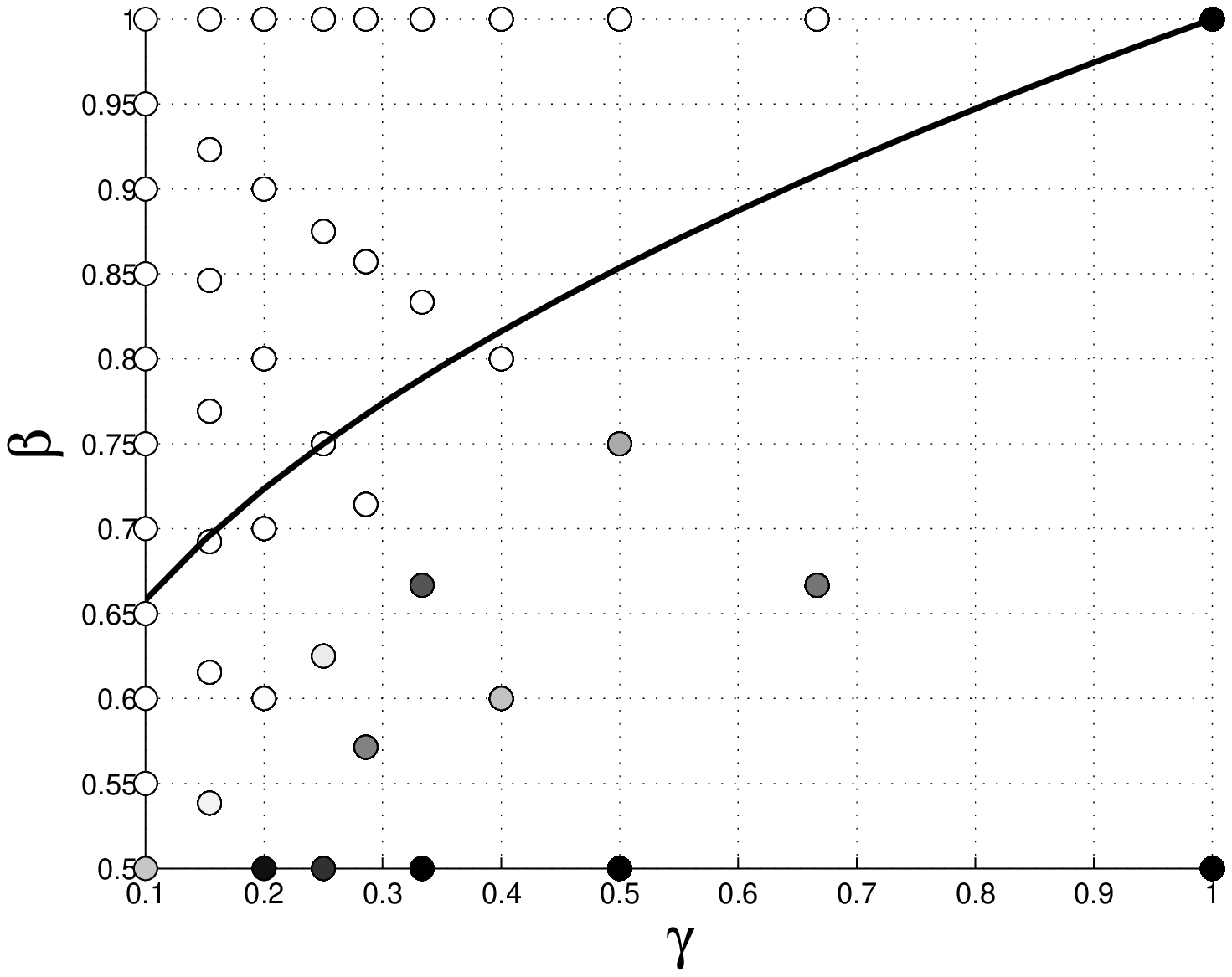}
\label{subfig:weakbound2}}\\
\subfigure[$n=20,~m=10$.]{
\includegraphics[width=0.4\linewidth]{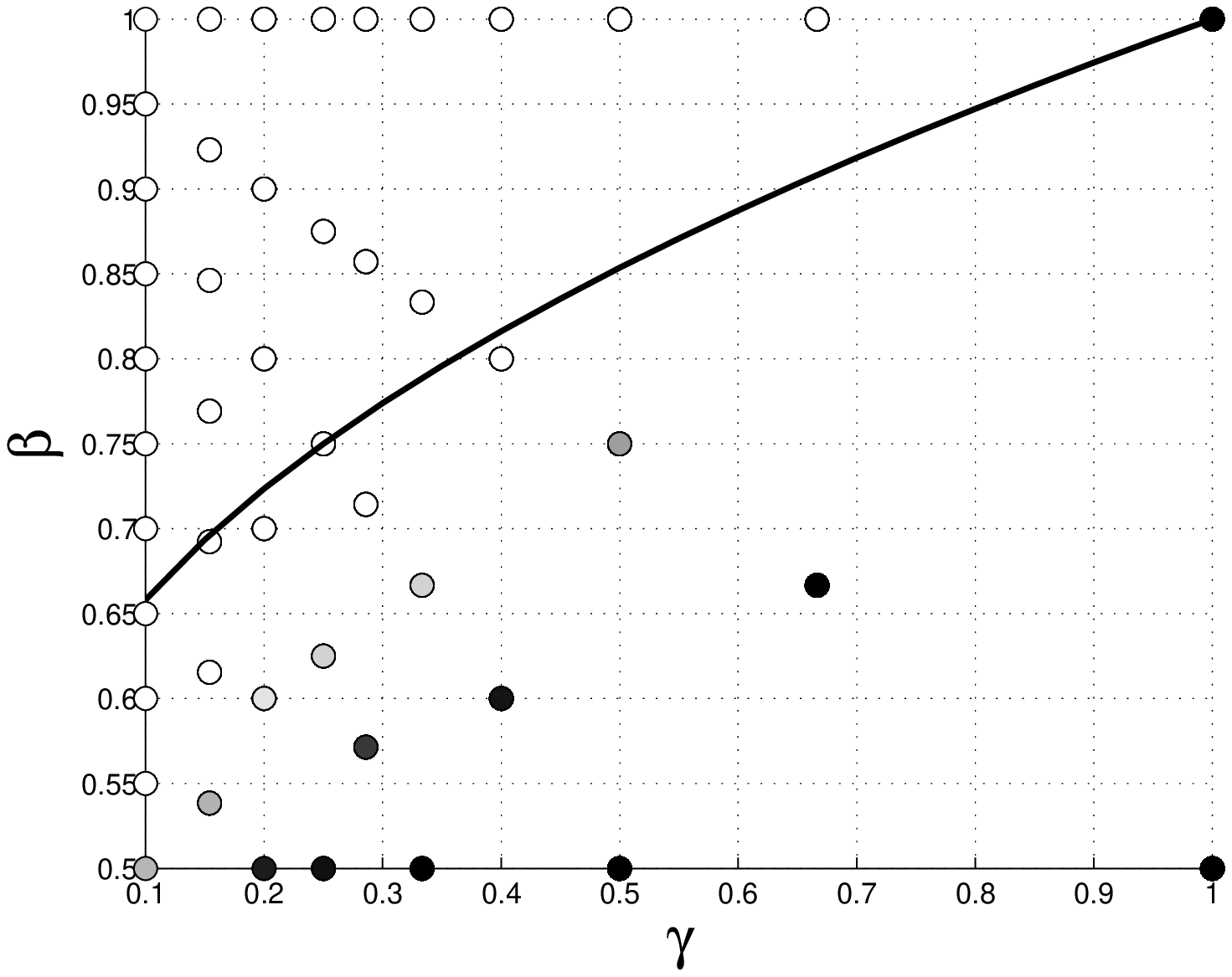}
\label{subfig:weakbound3}}
\subfigure[$n=20,~m=2$.]{
\includegraphics[width=0.4\linewidth]{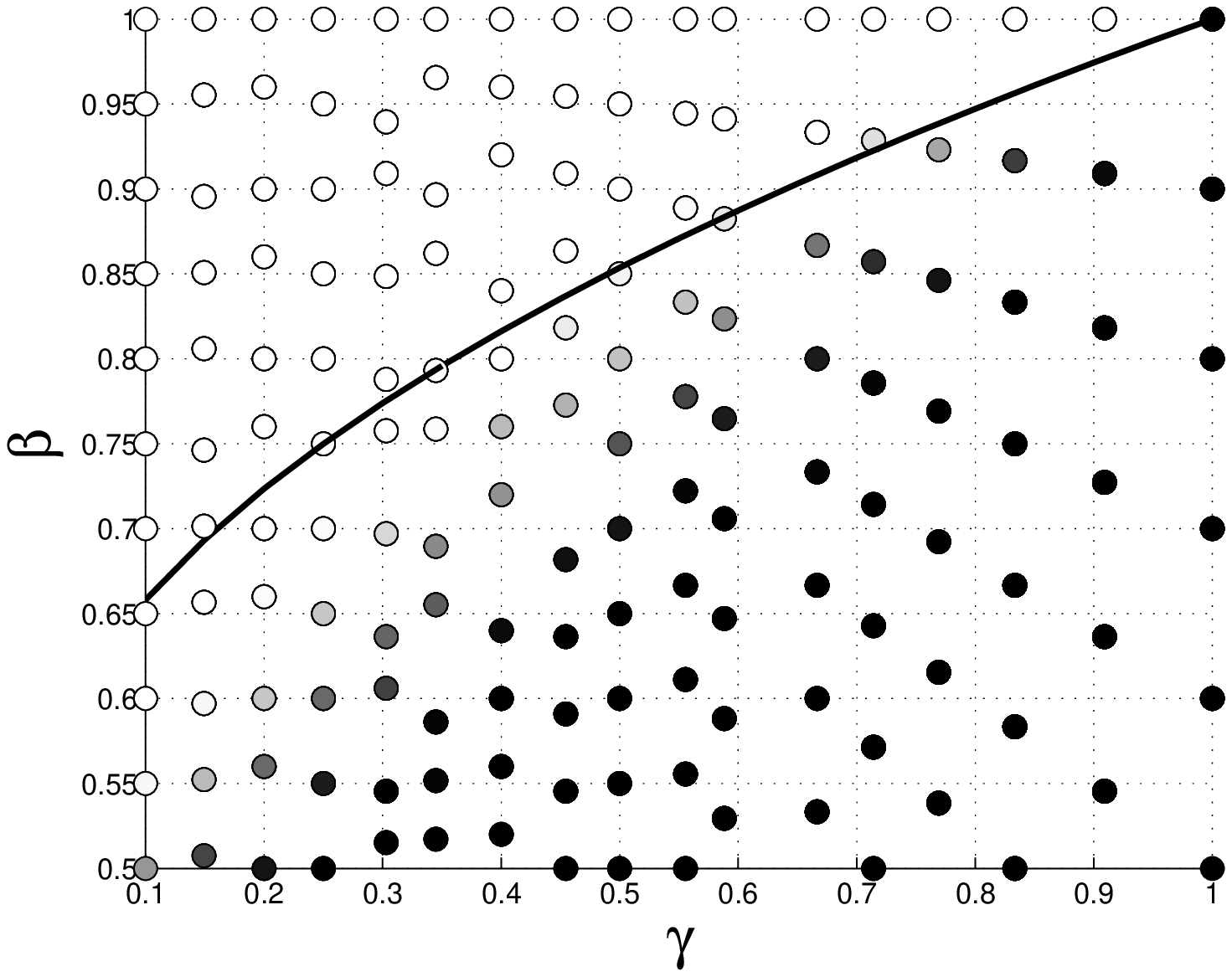}
\label{subfig:weakbound4}}
%\vspace*{-1em}
\caption{Empirical probability of success for \eqref{eq:P1} and the weak bound of \eqref{eq:weakbound} (solid black line). Empty circles correspond to quadruplets $(n,m,k,s)$ with perfect empirical recovery and solid black circles to problem setups having failed in all experiments.}
\label{fig:weakbound}
\end{figure}

The results are depicted in Fig.~\ref{fig:weakbound}. Every pair $(\gamma(k),\beta(s,k))$ corresponds to a circle whose face intensity indicates the probability of recovery as explained in the caption. The east and south-east parts of  Figs.~\ref{subfig:weakbound1}-\ref{subfig:weakbound3} are not as crowded, since for $\gamma$ close to 1, the integer $k$ becomes small, which implies that there are not many choices for an integer $s\in\left[k/2,{k}\right]$. The condition for highly probable recovery in the weak sense, $\beta=\left(\sqrt{\gamma}+1\right)/2$, is also shown as a black solid curve. According to the weak bound \eqref{eq:weakbound}, the circles above this curve correspond to dimension setups with high probability of success for large $n$. The empirically evaluated probabilities validate the result even for moderate values of $n$.

\subsection{Test Cases for RS}
The RS solvers developed are numerically compared in this subsection. The setup involves a network of $k=16$ sensors collecting observation vectors of size $m=4$, and an unknown vector of size $n=20$. Quantities $\mathbf{x}_0$, $\mathbf{A}$, and $\mathbf{b}$, all follow the model of the previous experiment, and the number of consistent sensors ranges from 8 to 16.

%%% \label{tbl:noiseless}

The comparison includes: (i) the LS solution of \eqref{eq:LS}; (ii) the $\ell_1$-error regression solution of \eqref{eq:L1}; (iii) the \eqref{eq:P1} solver; and (iv) the \eqref{eq:P2} solver obtained after one iteration of \eqref{eq:P2a}. In addition, a genie-aided LS (GA-LS) solver knowing a priori the reliable sensors, $\hat{\mathbf{x}}_{GA-LS}:=(\mathbf{A}^T_{\mathcal{S}_0} \mathbf{A}_{\mathcal{S}_0})^{-1}\mathbf{A}^T_{\mathcal{S}_0}\mathbf{b}_{\mathcal{S}_0}$, is implemented to serve as a benchmark. The parameter $\delta$ in \eqref{eq:P2a} is set to $10^{-4}$, whereas the simulation results were insensitive to the range of values from $10^{-2}$ to $10^{-8}$.

The sensor detection probability is empirically estimated through 1,000 Monte Carlo experiments. An estimate $\hat{\mathbf{x}}$ is considered to have successfully classified the sensors whenever the residual $\|\mathbf{b}_i-\mathbf{A}_i\hat{\mathbf{x}}\|_{\infty}$ is smaller than or equal to $10^{-4}$ for $i\in\mathcal{S}_0$, and larger than $10^{-4}$ for $i\in\bar{\mathcal{S}}_0$. As evidenced by Table \ref{tbl:noiseless}, the LS solution fails to identify the reliable subset. In contrast, the novel \eqref{eq:P1} scheme shows a clear advantage over the $\ell_1$-error regression solution, while the empirical detection probability further improves for the \eqref{eq:P2} method, even after a single iteration.

\subsection{Test Cases for RSN}
To evaluate the developed RSN solvers, the unknown vector was fixed at $\mathbf{x}_0=\mathbf{1}_n/\sqrt{n}$, while the reliable sensors followed the model $\mathbf{b}_{\mathcal{S}_0} = \mathbf{A}_{\mathcal{S}_0}\mathbf{x}_0+\mathbf{n}_{\mathcal{S}_0}$, with $\mathbf{n}_{\mathcal{S}_0}\sim \mathcal{N}(\mathbf{0}, \sigma^2 \mathbf{I}_{sm})$ and known $\sigma$. A plausible figure of merit in this scenario is the MSE, $\mathbb{E}[\|\mathbf{x}_0- \hat{\mathbf{x}}\|_2^2]$, which was empirically estimated by averaging over 1,000 Monte Carlo experiments.

%%% \label{fig:noisy_white}

Comparisons included: (i) the LS estimator; (ii) the GA-LS estimator; (iii) the $\ell_1$-error estimator of \eqref{eq:L1}; (iv) the conventional (scalar) Huber's M-estimator of \eqref{eq:P3}; (v) the \eqref{eq:P1} solver; (vi) the one-iteration solution of \eqref{eq:P2}; (vii) the \eqref{eq:P3} solver; and (viii) the one-iteration solution of \eqref{eq:P4}. The value of $\delta$ parameters in \eqref{eq:P2} and \eqref{eq:P4} turned out to be not critical, and were set to $10^{-4}$. The cutoff parameter $\tau$ for the Huber's M-estimator was selected as $1.34\sigma$, whereas $\lambda$ in both \eqref{eq:P3} and \eqref{eq:P4} was set to $1.34\sigma \sqrt{m}$. It is worth noting that the average number of iterations for the block-coordinate descent algorithm of Subsection \ref{subsec:bcd_algorithm} was between 16 (for SNR$=10$ dB) and 30 (for SNR$=25$ dB), while its execution time was 1,000 times lower than that of a standard SOCP solver.

\begin{figure}
\centering
\subfigure[White noise.]{
\includegraphics[width=0.50\linewidth]{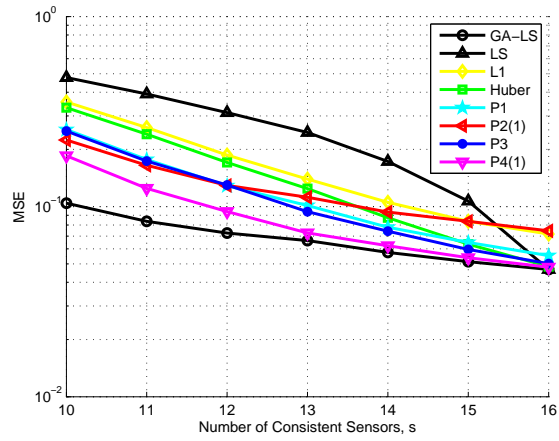}
\label{fig:noisy_white}}
\subfigure[Colored noise.]{
\includegraphics[width=0.50\linewidth]{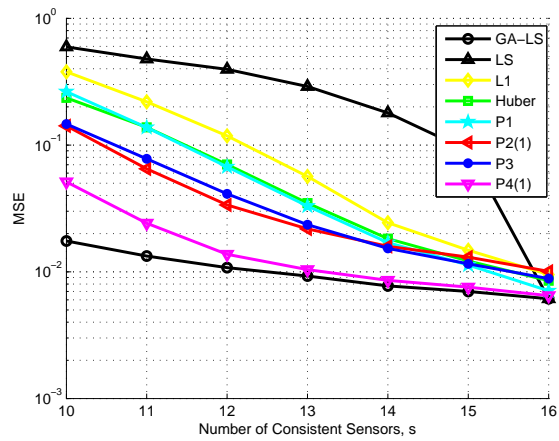}
\label{fig:noisy_colored}}
\caption{MSE performance for RSN with ($n,m,k)=(20,4,16)$.}
\end{figure}

In Fig.~\ref{fig:noisy_white}, the MSE achieved by each method is plotted versus the number of consistent sensors $s$ for SNR $=10$~dB. The curves show that the block-sparsity ignorant LS, $\ell_1$, and Huber's estimators are generally outperformed by the novel schemes. The \eqref{eq:P1} and \eqref{eq:P2} solvers, originally designed for the RS task, still exhibit reasonable performance that worsens as $s\rightarrow k$. The \eqref{eq:P3} estimator shows a slight improvement; but its solution serves as a good initialization point for the one-iteration estimates of \eqref{eq:P4}. Note that the derived RSN solvers combine robustness with efficiency in the absence of outliers.

%%% \label{fig:noisy_colored}

To test the effect of correlated sensor measurements, the following experiment was performed. The reliable sensors were modeled again as $\mathbf{b}_{\mathcal{S}_0} = \mathbf{A}_{\mathcal{S}_0}\mathbf{x}_0+\mathbf{n}_{\mathcal{S}_0}$, the unreliable ones as $\mathbf{b}_{\bar{\mathcal{S}}_0} = \mathbf{n}_w+\mathbf{n}_{\bar{\mathcal{S}_0}}$ where $\mathbf{n}_w\sim \mathcal{N}(\mathbf{0},\mathbf{I}_{(k-s)m})$, while $[\mathbf{n}_{\mathcal{S}_0}^T~\mathbf{n}_{\bar{\mathcal{S}}_0}^T]^T \sim \mathcal{N}(\mathbf{0},\mathbf{\Sigma})$ and $\mathbf{\Sigma}$ is a symmetric Toeplitz matrix with first column $[1~ 0.9~ 0.9^2~ \cdots~ 0.9^{km-1}]^T$. The two RSN solvers were modified according to Remark \ref{re:colored}. Fig.~\ref{fig:noisy_colored} shows the MSE curves obtained at SNR $= 10$~dB. In this correlated noise setup, the superiority of RSN solvers is even more prominent.

Correctly classifying the sensors as reliable/unreliable is critical. Once a method has completed this classification task, the estimation of $\mathbf{x}_0$ can be performed based solely on the sensors classified as reliable. Assuming successful classification, the MSE performance of GA-LS can be attained. The probability of correct sensor classification was evaluated in another simulation setup that differs from the previous ones in the following ways: problem dimensions were $(n,m,k)=(80,8,32)$; the reliable sensors followed the linear white Gaussian model at SNR $= 5$~dB; $\mathbf{b}_{\bar{\mathcal{S}}_0}$ had entries independently drawn from the zero mean Laplacian distribution with variance $(\sigma^2+1)$; and $\tau$ and $\lambda$ parameters were set to $\sigma$ and $\sigma\sqrt{m}$, respectively. The solvers (i)-(iii) and (iv)-(v) do not provide a classification mechanism, hence, a sensor was deemed reliable when its residual $\ell_2$-norm was smaller than $10^{-4}$. The Huber's estimator (iv) can identify outlying scalar measurements and a sensor was considered correctly classified when all its measurements were correctly classified. For $(P_3)$ and $(P_4)$, the identification followed naturally from the $\mathbf{u}_i$ vectors. The results are listed in Table~\ref{tbl:noisy}. The majority of methods fail to identify the reliable sensors and yield an empirical probability close to $(1-s/k)$, which is the ratio of unreliable sensors. The improvement offered by Huber's estimator is marginal, while ($P_3$) and in particular ($P_4$) outperform all others.

\section{Conclusions}\label{sec:conclusions}
Contemporary approaches to compressive sampling and variable selection in linear regression problems exploit (block) sparsity present in the signal of interest. The fresh perspective offered in this work broadens the scope of sparsity-exploiting algorithms to settings where model mismatch induced by unreliable sensors or outliers gives rise to (block) sparse residuals, even when the signal of interest is not sparse. This perspective links compressive sampling and sparse linear regression with two important problems: (i) finding the maximum number of feasible subsystems of linear equations; and (ii) robust multivariate linear regression. Capitalizing on these links, robust sensing algorithms were developed to reveal unreliable sensors and recover the signal of interest based on reliable sensors. In the absence of noise, necessary and sufficient conditions were provided for exact recovery (identifiability). Their probabilistic characterization showed that they hold with overwhelming probability when the regression matrix is Gaussian distributed. In the presence of noise, the RS task was reformulated to a combinatorial problem that was subsequently surrogated by (non-)convex costs. The two subsystem-aware robust estimators derived can be solved by an efficient block coordinate descent algorithm. The simulated tests demonstrated that all proposed schemes succeed in the task for which they have been designed for.

\appendix
\begin{IEEEproof}[Proof of Theorem \ref{th:rsconditions}]
The sufficiency of the conditions in \eqref{eq:rsconditions} is shown first. Recall from Lemma \ref{le:unique-exact} that the conditions in \eqref{eq:rsconditions} imply that $\mathbf{x}_0$ is the unique minimizer of \eqref{eq:P0}. Let $\mathcal{S}$ denote the set of reliable wrt $\mathbf{x}_0$ sensors with $|\mathcal{S}|=s>k/2$ for which $\mathbf{b}_{\mathcal{S}}= \mathbf{A}_{\mathcal{S}} \mathbf{x}_0$. Vector $\mathbf{x}_0$ is the unique minimizer of \eqref{eq:P1} too if and only if the vector $(\mathbf{x}_0-\mathbf{u})$ for any nonzero $\mathbf{u}\in \mathbb{R}^n$ yields a strictly larger \eqref{eq:P1} cost than $\mathbf{x}_0$ does. Indeed, letting $\mathbf{v}:=\mathbf{A}\mathbf{u}$, the cost attained by $(\mathbf{x}_0-\mathbf{u})$ is 
\begin{align*}
&\sum_{i=1}^k\|\mathbf{b}_{i}-\mathbf{A}_i\mathbf{x}_0+\mathbf{v}_i\|_2 = 
\sum_{i\in \mathcal{S}}\|\mathbf{b}_{i}-\mathbf{A}_i\mathbf{x}_0+\mathbf{v}_i\|_2 +
\sum_{i\in \bar{\mathcal{S}}}\|\mathbf{b}_{i}-\mathbf{A}_i\mathbf{x}_0+\mathbf{v}_i\|_2\\
& \overset{(a)}{=} \sum_{i\in \mathcal{S}}\|\mathbf{v}_i\|_2 
+ \sum_{i\in \bar{\mathcal{S}}}\|\mathbf{b}_{i}-\mathbf{A}_i\mathbf{x}_0+\mathbf{v}_i\|_2\\
&\overset{(b)}{\geq} \sum_{i\in \mathcal{S}}\|\mathbf{v}_i\|_2 +
\sum_{i\in \bar{\mathcal{S}}}\|\mathbf{b}_{i}-\mathbf{A}_i\mathbf{x}_0\|_2 -
\sum_{i\in \bar{\mathcal{S}}}\|\mathbf{v}_i\|_2 \overset{(c)}{>}
\sum_{i=1}^k\|\mathbf{b}_{i}-\mathbf{A}_i\mathbf{x}_0\|_2
\end{align*}
where equality $(a)$ uses that $\mathbf{b}_{\mathcal{S}}= \mathbf{A}_{\mathcal{S}} \mathbf{x}_0$, inequality $(b)$ stems from the reverse triangle inequality, and inequality $(c)$ is due to the assumed conditions of the theorem, and again the fact that $\mathbf{b}_{\mathcal{S}}= \mathbf{A}_{\mathcal{S}} \mathbf{x}_0$.

Necessity is shown by proving the contrapositive. Specifically, it must be shown that if there exists a $\mathbf{v}\in \range(\mathbf{A})$ and an $(\mathcal{S},\bar{\mathcal{S}})$ partition of $\mathcal{I}$ with $|\mathcal{S}|=s$ for which $\sum_{i\in \mathcal{S}}\|\mathbf{v}_i\|_2 \leq \sum_{i\in \bar{\mathcal{S}}}\|\mathbf{v}_i\|_2$, then there exists an $\mathbf{x}_0$ that attains a minimum \eqref{eq:P0} cost of $s$, but is not the unique minimizer of \eqref{eq:P1}. Suppose that $\mathbf{b}_{\mathcal{S}}:=\mathbf{A}_{\mathcal{S}}\mathbf{x}_0$ and $\mathbf{b}_{\bar{\mathcal{S}}}:=\mathbf{A}_{\bar{\mathcal{S}}}\mathbf{x}_0/2$ for an $(\mathcal{S},\bar{\mathcal{S}})$ partition with $|\mathcal{S}|=s>k/2$. Vector $\mathbf{x}_0$ obviously minimizes \eqref{eq:P0}, whereas $\mathbf{x}_0/2$ does not since $|\bar{\mathcal{S}}|< |\mathcal{S}|$. Assume $\mathbf{v}:=\mathbf{A}\mathbf{x}_0\in \range(\mathbf{A})$ and $\sum_{i\in \mathcal{S}}\|\mathbf{v}_i\|_2 \leq \sum_{i\in \bar{\mathcal{S}}}\|\mathbf{v}_i\|_2$. It is easy to check that the \eqref{eq:P1} costs attained by $\mathbf{x}_0/2$ and $\mathbf{x}_0$ are respectively $\sum_{i\in \mathcal{S}}\|\mathbf{v}_i\|_2/2$ and $\sum_{i\in \bar{\mathcal{S}}}\|\mathbf{v}_i\|_2/2$. Hence, it has been shown that $\mathbf{x}_0/2$ attains a \eqref{eq:P1} cost not greater than that of $\mathbf{x}_0$, i.e., $\mathbf{x}_0$ is not the unique minimizer of \eqref{eq:P1}. This concludes the proof. 
\end{IEEEproof}

\begin{lemma}[Lipschitz continuity of $f(\mathbf{A})$]\label{le:lipschitz}
The function $f(\mathbf{A})$ defined in \eqref{eq:fA} is Lipschitz continuous with Lipschitz constant at most $\sqrt{k}$.
\end{lemma}

\begin{IEEEproof}
Let $\mathbf{A},~\mathbf{A}'\in\mathbb{R}^{km\times n}$ and $\mathbf{w},~\mathbf{w}'\in\mathbb{R}^n$ be the minimizing arguments of $f(\mathbf{A})$ and $f(\mathbf{A}')$, respectively. The difference of the function at these two points is
\begin{align*}
f(\mathbf{A})-f(\mathbf{A}')&= \left(\sum_{i\in \mathcal{S}} \|\mathbf{A}_i\mathbf{w}\|_2 - \sum_{i\in \bar{\mathcal{S}}} \|\mathbf{A}_i\mathbf{w}\|_2\right) - 
\left(\sum_{i\in \mathcal{S}} \|\mathbf{A}_i'\mathbf{w}'\|_2 - \sum_{i\in \bar{\mathcal{S}}} \|\mathbf{A}_i'\mathbf{w}'\|_2\right)\\
&\overset{(a)}{\leq} \left(\sum_{i\in \mathcal{S}} \|\mathbf{A}_i\mathbf{w}'\|_2 - \sum_{i\in \bar{\mathcal{S}}} \|\mathbf{A}_i\mathbf{w}'\|_2\right)-
\left(\sum_{i\in \mathcal{S}} \|\mathbf{A}_i'\mathbf{w}'\|_2 - \sum_{i\in \bar{\mathcal{S}}} \|\mathbf{A}_i'\mathbf{w}'\|_2\right)\\
&\overset{(b)}{\leq} \sum_{i\in \mathcal{S}} \|(\mathbf{A}_i-\mathbf{A}_i')\mathbf{w}'\|_2 + \sum_{i\in \bar{\mathcal{S}}} \|(\mathbf{A}_i-\mathbf{A}_i')\mathbf{w}'\|_2
\overset{(c)}{\leq} \sup_{\|\mathbf{u}\|_2=1} \sum_{i=1}^k \|\tilde{\mathbf{A}}_i\mathbf{u}\|_2
\end{align*}
where inequality $(a)$ holds because $\mathbf{w}$ is by definition the minimizer of $f(\mathbf{A})$; $(b)$ follows from the reverse triangle inequality applied on each subset; $(c)$ holds trivially for $\|\mathbf{w}'\|_2=1$; and $\tilde{\mathbf{A}}_i:=\mathbf{A}_i-\mathbf{A}_i'$.

Now, define the function appearing in the right-hand side of the last inequality as
\begin{equation}\label{eq:gfunction2}
g(\tilde{\mathbf{A}}) := \sup_{\|\mathbf{u}\|_2=1} \sum_{i=1}^k \|\tilde{\mathbf{A}}_i\mathbf{u}\|_2
\end{equation}
so that $f(\mathbf{A})-f(\mathbf{A}')\leq g(\tilde{\mathbf{A}})$. Since $f(\mathbf{A}')-f(\mathbf{A})\leq g(-\tilde{\mathbf{A}})=g(\tilde{\mathbf{A}})$, it holds that
$\left|f(\mathbf{A})-f(\mathbf{A}')\right|\leq g(\tilde{\mathbf{A}})$. Given that $g(\mathbf{0})=0$, if $g(\mathbf{A})$ is Lipschitz continuous with constant at most $L$, i.e., $|g(\mathbf{A})|\leq L \|\mathbf{A}\|_F$, where $\|\mathbf{A}\|_F$ is the Frobenius norm of matrix $\mathbf{A}$, then $f(\mathbf{A})$ is also Lipschitz continuous and its constant is at most $L$. Hence, it suffices to show that $g(\mathbf{A})$ is Lipschitz continuous and its constant is upper bounded by $\sqrt{k}$.

To proceed, recall first that the $\ell_2$-norm of a vector $\mathbf{x}\in\mathbb{R}^n$ can be written as \cite[p.~637]{BoVa04}
\begin{equation}\label{eq:norm}
\|\mathbf{x}\|_2=\sup\{ \mathbf{x}^T\mathbf{y}:\|\mathbf{y}\|_2\leq 1\}.
\end{equation}
Using \eqref{eq:norm}, $g(\mathbf{A})$ can alternatively be expressed as
\begin{equation}\label{eq:gfunction}
g(\mathbf{A})=\sup_{\|\mathbf{u} \|_2=1}\sup_{\|\mathbf{v}_i\|_2\leq 1}\sum_{i=1}^k \mathbf{v}_i^T\mathbf{A}_i\mathbf{u}
\end{equation}
which is a supremum over infinitely many linear functions of $\mathbf{A}$, and as such it is convex. Recall that if a function $f:\mathbb{R}^p\rightarrow \mathbb{R}$ is convex with a subgradient $s(\mathbf{x})$ for which $\|\sup_{\mathbf{x}}s(\mathbf{x})\|_2$ is finite, then $f$ is Lipschitz with constant $L\leq \|\sup_{\mathbf{x}}s(\mathbf{x})\|_2$. This claim can be proved by the definitions of the subgradient and the Lipschitz constant. Thus, it suffices to find a subgradient of $g(\mathbf{A})$ and upper bound its norm.

If $\mathbf{u}_{\ast}$ and $\{\mathbf{v}_{\ast,i}\}_{i=1}^k$ are the maximizers of $g(\mathbf{A})$, then a subgradient is given by the matrix $\mathbf{G}(\mathbf{A})=\left[\mathbf{u}_{\ast}\mathbf{v}_{\ast,1}^T ~ \cdots ~ \mathbf{u}_{\ast}\mathbf{v}_{\ast,k}^T \right]$ with $\|\mathbf{u_{\ast}}\|_2=1$ and $\|\mathbf{v_{\ast,i}}\|_2\leq 1$ for $i=1,\ldots,k$. The norm of this subgradient is
\begin{equation*}
\|\mathbf{G}(\mathbf{A})\|_F=
\sqrt{\sum_{i=1}^k\|\mathbf{u}_{\ast}\mathbf{v}_{\ast,i}^T\|_F^2}= 
\sqrt{\sum_{i=1}^k\|\mathbf{u}_{\ast}\|_2^2\|\mathbf{v}_{\ast,i}\|_2^2}=
\sqrt{\sum_{i=1}^k\|\mathbf{v}_{\ast,i}\|_2^2}\leq
\sqrt{k}.
\end{equation*}
The bound is independent of $\mathbf{A}$, and the proof is complete.
\end{IEEEproof}

\begin{lemma}[Expected value lower bound]\label{le:meanlb}
For the random matrix $\mathbf{A}\in\mathbb{R}^{km\times n}$ with entries drawn independently from $\mathcal{N}\left(0,1\right)$, it holds that $\mathbb{E}\left[f(\mathbf{A})\right]\geq \mu$ with
$\mu:= \left(\frac{2\beta-1}{\sqrt{\gamma}}-1\right)\sqrt{kn}\left(1+o_n(1)\right)$.
\end{lemma}

\begin{IEEEproof}
Consider rewriting $f(\mathbf{A})$ using \eqref{eq:norm} as
\begin{equation}\label{eq:fA2}
f(\mathbf{A})=\inf_{\|\mathbf{u}\|_2=1} \sup_{\|\mathbf{v}_i\|_2\leq 1} \inf_{\|\mathbf{z}_i\|_2\leq1}
\sum_{i\in \mathcal{S}}\mathbf{v}_i^T\mathbf{A}_i\mathbf{u} + \sum_{i\in \bar{\mathcal{S}}}\mathbf{z}_i^T\mathbf{A}_i\mathbf{u}.
\end{equation}

Next, introduce auxiliary random vectors $\mathbf{y}\in\mathbb{R}^n$, $\mathbf{s}_i\in\mathbb{R}^m$, $\mathbf{t}_i\in\mathbb{R}^m$ for $i=1,\ldots,k$, and $w\in \mathbb{R}$ having their entries drawn independently from $\mathcal{N}(0,1)$, and define the functionals
\begin{subequations}
\begin{align}
h_y\left(\mathbf{u},\mathbf{v}_i,\mathbf{z}_i\right)&:=\sum_{i\in \mathcal{S}}\mathbf{v}_i^T\mathbf{A}_i\mathbf{u}
+ \sum_{i\in \bar{\mathcal{S}}}\mathbf{z}_i^T\mathbf{A}_i\mathbf{u} + \sqrt{k}w\label{eq:hfun}\\
h_x\left(\mathbf{u},\mathbf{v}_i,\mathbf{z}_i\right)
&:=\sum_{i\in \mathcal{S}}\mathbf{v}_i^T\mathbf{s}_i
+ \sum_{i\in \bar{\mathcal{S}}}\mathbf{z}_i^T\mathbf{t}_i + \sqrt{k}\mathbf{u}^T\mathbf{y},~\textrm{for}\label{eq:gfun}
\end{align}
\end{subequations}
\begin{equation}\label{eq:uvzconditions}
\|\mathbf{u}\|_2=1,~ \{\|\mathbf{v}_i\|_2\leq1\}_{i=1}^k~\textrm{and}~ \{\|\mathbf{z}_i\|_2\leq1\}_{i=1}^k.
\end{equation}

Consider now the triplets $\left(\mathbf{u},\mathbf{v}_i,\mathbf{z}_i\right)$ and $\left(\mathbf{u}',\mathbf{v}_i',\mathbf{z}_i'\right)$. By using the i.i.d. property of the random variables appearing in the functionals, it holds that
\begin{align*}
\mathbb{E}\left[h_y\left(\mathbf{u},\mathbf{v}_i,\mathbf{z}_i\right) h_y\left(\mathbf{u}',\mathbf{v}_i',\mathbf{z}_i'\right)\right] &= 
\mathbf{u}^T\mathbf{u}'\left(\sum_{i\in\mathcal{S}}\mathbf{v}_i^T\mathbf{v}_i' + \sum_{i\in\bar{\mathcal{S}}}\mathbf{z}_i^T\mathbf{z}_i'\right) + k\\
\mathbb{E}\left[h_x\left(\mathbf{u},\mathbf{v}_i,\mathbf{z}_i\right) h_x\left(\mathbf{u}',\mathbf{v}_i',\mathbf{z}_i'\right)\right] &= \sum_{i\in\mathcal{S}} \mathbf{v}_i^T\mathbf{v}_i' + \sum_{i\in\bar{\mathcal{S}}}\mathbf{z}_i^T\mathbf{z}_i' + k\mathbf{u}^T\mathbf{u}'
\end{align*}
whereas the difference of the two expectations is
\begin{equation*}\label{eq:diff}
\mathbb{E}\left[h_y\left(\mathbf{u},\mathbf{v}_i,\mathbf{z}_i\right) h_y\left(\mathbf{u}',\mathbf{v}_i',\mathbf{z}_i'\right)\right] {-} \mathbb{E}\left[h_x\left(\mathbf{u},\mathbf{v}_i,\mathbf{z}_i\right) h_x\left(\mathbf{u}',\mathbf{v}_i',\mathbf{z}_i'\right)\right] = 
\left(\mathbf{u}^T\mathbf{u}'{-}1\right) \left(\sum_{i\in\mathcal{S}} \mathbf{v}_i^T\mathbf{v}_i' + \sum_{i\in\bar{\mathcal{S}}}\mathbf{z}_i^T\mathbf{z}_i' - k\right).
\end{equation*}
By exploiting the properties of vectors $\mathbf{u}$, $\mathbf{v}_i$, and $\mathbf{z}_i$ in \eqref{eq:uvzconditions}, it follows readily that
\begin{subequations}\label{eq:ineq}
\begin{align}
\mathbb{E}\left[h_x\left(\mathbf{u},\mathbf{v}_i,\mathbf{z}_i\right) h_x\left(\mathbf{u},\mathbf{v}_i',\mathbf{z}_i'\right)\right] &=
\mathbb{E}\left[h_y\left(\mathbf{u},\mathbf{v}_i,\mathbf{z}_i\right) h_y\left(\mathbf{u},\mathbf{v}_i',\mathbf{z}_i'\right)\right],\label{eq:ineq_a}\\ 
\mathbb{E}\left[h_x\left(\mathbf{u},\mathbf{v}_i,\mathbf{z}_i\right) h_x\left(\mathbf{u}',\mathbf{v}_i',\mathbf{z}_i'\right)\right] &\leq \mathbb{E}\left[h_y\left(\mathbf{u},\mathbf{v}_i,\mathbf{z}_i\right) h_y\left(\mathbf{u}',\mathbf{v}_i',\mathbf{z}_i'\right)\right].\label{eq:ineq_b}
\end{align}
\end{subequations}
To proceed, the following lemma is needed \cite[Cor.~10]{Go88}.

\begin{lemma}[\cite{Go88}]\label{le:gordon}
Let $\{X_{ijk}\}$ and $\{Y_{ijk}\}$ be two zero-mean Gaussian processes indexed by $(i,j,k)$ for $i=1,\ldots,n$, $j=1,\ldots,m$, and $k=1,\ldots,p$, which satisfy the following conditions:
\begin{description}
\item[(c1)] $\mathbb{E}\left[X_{ijk}^2\right]=\mathbb{E}\left[Y_{ijk}^2\right]$ for all $(i,j,k)$.
\item[(c2)] For any two triplets $\alpha=(i,j,k)$ and $\alpha'=(i',j',k')$, $\mathbb{E}\left[X_{\alpha}X_{\alpha'}\right]\geq \mathbb{E}\left[Y_{\alpha}Y_{\alpha'}\right]$ if $i=i'$ and $j\neq j'$, and $\mathbb{E}\left[X_{\alpha}X_{\alpha'}\right]\leq \mathbb{E}\left[Y_{\alpha}Y_{\alpha'}\right]$ in all other cases.
\end{description}
Under (c1) and (c2), it holds that 
\begin{equation}\label{eq:gordon}
\mathbb{E}\left[\max_i\min_j\max_k X_{ijk}\right] \geq 
\mathbb{E}\left[\max_i\min_j\max_k Y_{ijk}\right].
\end{equation}
\end{lemma}

Even though the indexes $(i,j,k)$ are denumerable, by using the compactness argument of \cite[Pr.~1]{ReXuHa11}, the comparison in \eqref{eq:gordon} extends to minimizations/maximizations over compact sets as well. Mapping the $X_{ijk}$ ($Y_{ijk}$) variables of Lemma \ref{le:gordon} to $-h_x(\mathbf{u},\mathbf{v}_i,\mathbf{z}_i)$ ($-h_y(\mathbf{u},\mathbf{v}_i,\mathbf{z}_i)$), it can be verified that the conditions of the lemma are met (cf. \eqref{eq:ineq}), and upon using \eqref{eq:gordon} deduce that
\begin{align*}
\mathbb{E} \left[ \sup_{\|\mathbf{u}\|_2=1}\inf_{\|\mathbf{z}_i\|_2\leq 1} \sup_{\|\mathbf{v}_i\|_2\leq 1}
-h_x\left(\mathbf{u},\mathbf{v}_i,\mathbf{z}_i\right)\right] &\geq
\mathbb{E} \left[ \sup_{\|\mathbf{u}\|_2=1}\inf_{\|\mathbf{z}_i\|_2\leq 1} \sup_{\|\mathbf{v}_i\|_2\leq 1}
-h_y\left(\mathbf{u},\mathbf{v}_i,\mathbf{z}_i\right)\right].
\end{align*}
Given that $\sup_x-f(x)=-\inf_{x}f(x)$, the previous inequality is equivalent to
\begin{align*}
\mathbb{E} \left[\inf_{\|\mathbf{u}\|_2=1}\inf_{\|\mathbf{z}_i\|_2\leq 1} \sup_{\|\mathbf{v}_i\|_2\leq 1}
h_x\left(\mathbf{u},\mathbf{v}_i,\mathbf{z}_i\right)\right] &\leq
\mathbb{E}\left[ \inf_{\|\mathbf{u}\|_2=1}\inf_{\|\mathbf{z}_i\|_2\leq 1} \sup_{\|\mathbf{v}_i\|_2\leq 1}
h_y\left(\mathbf{u},\mathbf{v}_i,\mathbf{z}_i\right)\right].
\end{align*}
But since the random variable $w$ in \eqref{eq:hfun} is zero mean, the right-hand side of the last inequality is equal to the desired expected value, $\mathbb{E}\left[f\left(\mathbf{A}\right)\right]$. Thus, it has been established that
\begin{equation*}
\mathbb{E}\left[ f\left(\mathbf{A}\right)\right]\geq
\mathbb{E} \left[\inf_{\|\mathbf{u}\|_2=1}\inf_{\|\mathbf{z}_i\|_2\leq 1} \sup_{\|\mathbf{v}_i\|_2\leq 1}
h_x\left(\mathbf{u},\mathbf{v}_i,\mathbf{z}_i\right)\right].
\end{equation*}
Using the definition of $h_x\left(\mathbf{u}, \mathbf{v}_i,\mathbf{z}_i\right)$ and exploiting the separability of the optimization, as well as the properties in \eqref{eq:uvzconditions}, one arrives at
\begin{equation}\label{eq:inequality}
\mathbb{E}\left[ f\left(\mathbf{A}\right)\right]\geq 
s\mathbb{E}\left[\|\mathbf{s}_i\|_2\right] - (k-s)\mathbb{E}\left[\|\mathbf{t}_i\|_2\right] -\sqrt{k} \mathbb{E}\left[\|\mathbf{y}\|_2\right].
\end{equation}
Recall that if $\mathbf{x}\sim \mathcal{N}(\mathbf{0}_n,\mathbf{I}_n)$, then $\|\mathbf{x}\|_2$ is chi-distributed with $n$ degrees of freedom, and mean value 
\begin{equation}\label{eq:meanChi}
\mathbb{E}\left[\|\mathbf{x}\|_2\right]=\frac{\sqrt{2\pi}}{\Beta\left(\frac{n}{2},\frac{1}{2}\right)}
\end{equation}
where $\Beta\left(\cdot,\cdot\right)$ denotes the Beta function. Applying \eqref{eq:meanChi} three times in \eqref{eq:inequality} yields
\begin{equation*}
\mathbb{E}\left[ f\left(\mathbf{A}\right)\right]\geq \mu=(2s-k)\frac{\sqrt{2\pi}}{\Beta\left(\frac{m}{2},\frac{1}{2}\right)} - 
\sqrt{k} \frac{\sqrt{2\pi}}{\Beta\left(\frac{n}{2},\frac{1}{2}\right)}.
\end{equation*}
Using the standard approximation $\frac{\sqrt{2\pi}}{\Beta\left(\frac{n}{2},\frac{1}{2}\right)} = \sqrt{n}\big(1+o_n(1)\big)$ \cite[Formulas 6.1.46 and 6.2.2]{AbSt72}, and for fixed $\gamma=n/(km)$ and $k$, it also holds that $\frac{\sqrt{2\pi}}{\Beta\left(\frac{m}{2},\frac{1}{2}\right)} = \sqrt{n}/\sqrt{\gamma k}\big(1+o_n(1)\big)$. Thus, the bound $\mu$ can be compactly expressed as $\mu = \left(\frac{2\beta-1}{\sqrt{\gamma}}-1\right)\sqrt{kn}\left(1+o_n(1)\right)$, which concludes the proof.
\end{IEEEproof}

\bibliographystyle{IEEEtranS}
\bibliography{IEEEabrv,mnfls}

\begin{table}[!ht]
\caption{Empirical probability of successful sensor classification (\%).}
\label{tbl:tables}
%\vspace*{-2em}
\centering
\subtable[RS task with $(n,m,k)=(20,4,16)$.]{
\small
\centering
\begin{tabular}{|l|r|r|r|r|r|}
\hline
 & \multicolumn{5}{|c|}{\textbf{Number of consistent sensors $s$}}\\
\hline
\textbf{Method} & \textbf{8} & \textbf{10} & \textbf{12} & \textbf{14} & \textbf{16}\\
\hline
GA-LS			& 100.0	& 100.0 & 100.0 & 100.0 & 100.0\\
LS				& 50.0	& 37.5	& 25.0	& 12.5	& 100.0\\
$l_1$			& 51.4	& 46.3	& 94.6	& 100.0	& 100.0\\
$P_1$			& 53.5	& 67.4	& 99.6	& 100.0	& 100.0\\
$P_2(1)$	& 81.5	& 99.3	& 100.0	& 100.0	& 100.0\\
\hline
\end{tabular}
\label{tbl:noiseless}
}
\subtable[RSN task with $(n,m,k)=(80,8,32)$.]{
\small
\centering
\begin{tabular}{|l|r|r|r|r|r|}
\hline
 & \multicolumn{5}{|c|}{\textbf{Number of consistent sensors $s$}}\\
\hline
\textbf{Method} & \textbf{16} & \textbf{20} & \textbf{24} & \textbf{28} & \textbf{32}\\
\hline
GA-LS			&	50.0 &	37.5	&	25.0	&	12.5	&	0.0\\
LS				& 50.0	&	37.5	&	25.0	&	12.5	&	0.0\\
$l_1$			& 50.0	&	37.5	&	25.0	&	12.5	&	0.0\\
Huber's		& 53.2	&	43.9	& 36.0	& 27.9	& 20.1\\
$P_1$			& 50.1	&	37.6	&	25.1	&	12.6	&	0.1\\
$P_2(1)$	& 55.0	&	44.1	&	31.8	&	18.5	&	5.3\\
$P_3$			& 68.7	&	73.9	&	79.6	&	83.5	&	84.4\\
$P_4(1)$	& 72.6	&	82.8	&	90.7	&	96.1	&	99.1\\
\hline
\end{tabular}\label{tbl:noisy}
}
\end{table}

%\begin{figure}
%\centering
%\includegraphics[width=0.70\linewidth]{noisy_white.eps}
%%\vspace*{-2em}
%\caption{MSE performance for RSN with ($n,m,k)=(20,4,16)$.}\label{fig:noisy_white}
%\end{figure}
%
%\begin{figure}
%\centering
%\includegraphics[width=0.70\linewidth]{noisy_colored.eps}
%%\vspace*{-2em}
%\caption{MSE performance for RSN with colored noisy measurements with ($n,m,k)=(20,4,16)$.}\label{fig:noisy_colored}
%\end{figure}

\end{document}